\documentclass{article} 
\usepackage{preprint,times}

\usepackage{amsmath,amsfonts,bm}

\def\eqref#1{equation~\ref{#1}}

\def\1{\bm{1}}

\DeclareMathAlphabet{\mathsfit}{\encodingdefault}{\sfdefault}{m}{sl}
\SetMathAlphabet{\mathsfit}{bold}{\encodingdefault}{\sfdefault}{bx}{n}

\usepackage{multirow}
\usepackage{hyperref}
\usepackage{url}
\usepackage{booktabs}
\usepackage{graphicx}
\usepackage{booktabs}
\usepackage{adjustbox}

\title{Can Image Agents Complete Visual Commissions? A Benchmark and Modular Framework}

\author{
\textbf{
Tong Li\textsuperscript{1},
Lishuai Gao\textsuperscript{1,$\ddagger$},
Qi Zeng\textsuperscript{1},
Yifan Li\textsuperscript{1,2},
Cong Wei\textsuperscript{1},
Zhuoliang Kang\textsuperscript{1},
Xiaoming Wei\textsuperscript{1},
} \\
 \textsuperscript{1}Meituan \quad
\textsuperscript{2}University of Science and Technology of China\\
\textsuperscript{$\ddagger$}Project lead \\
\texttt{lt.litong@qq.com}
}

\arxivcopy 
\renewcommand{\headrulewidth}{0pt}
\renewcommand{\footrulewidth}{0pt}
\AddToShipoutPicture{%
  \AtPageLowerLeft{%
    \raisebox{0.34in}{\makebox[\paperwidth][c]{\thepage}}%
  }%
}

\usepackage{amssymb} 
\usepackage{graphicx}
\usepackage{booktabs}
\usepackage{titletoc}
\usepackage{tabularx}

\usepackage{cleveref}
\crefname{figure}{Figure}{Figures}
\crefname{table}{Table}{Tables}
\crefname{section}{Sec.}{Secs.}
\Crefname{section}{Section}{Sections}
\crefname{equation}{Eq.}{Eqs.}

\begin{document}

\maketitle

\begin{abstract}
Visual commissions require image agents to integrate heterogeneous source materials, preserve requirements across dependent stages, and produce coordinated deliverables. Existing evaluations examine specific capabilities and execution settings, leaving end-to-end commission completion less understood. We introduce Visual Commission Bench (VCB), comprising 600 tasks across 12 categories and three complementary tiers, from capability-focused tasks to complete commissions. VCB uses 5,039 requirement checks and separate visual quality ratings to assess content correctness, visual realization, constraint preservation, and delivery completeness without prescribing an internal workflow. We present V-Protocol, a reference agent that dynamically composes reusable task, cognitive, and expression modules at inference time. Task modules organize delivery, cognitive modules derive content, and expression modules guide visual realization. The modules exchange structured intermediate artifacts, while a shared execution harness tracks artifact revisions and dependencies, manages tool execution, and checks completion. With the same controller and image generator as the agent baselines, V-Protocol achieves 69.0\% strict task success on VCB, compared with 42.0\% for SCOPE. On Mind-Bench, strict accuracy increases from 39.2\% to 45.4\%. 
The gallery for V-protocol could be found in Figure~\ref{fig:gallery}.
\end{abstract}

\section{Introduction}
\label{sec:introduction}

\begin{figure*}[t]
\centering
\includegraphics[width=\linewidth]
{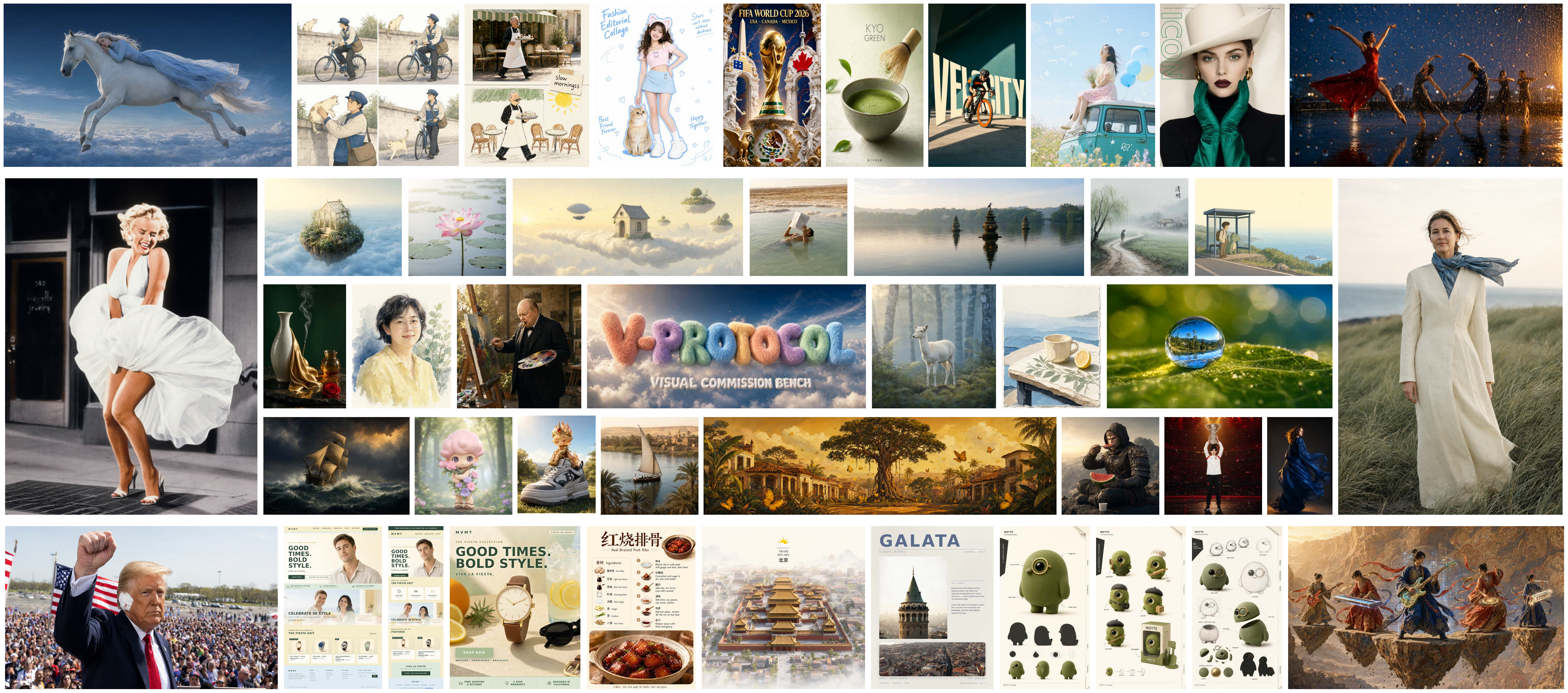}
\caption{\textbf{Gallery for V-protocol.}
}
\label{fig:gallery}
\end{figure*}

\begin{figure*}[t]
\centering
\includegraphics[width=\linewidth]{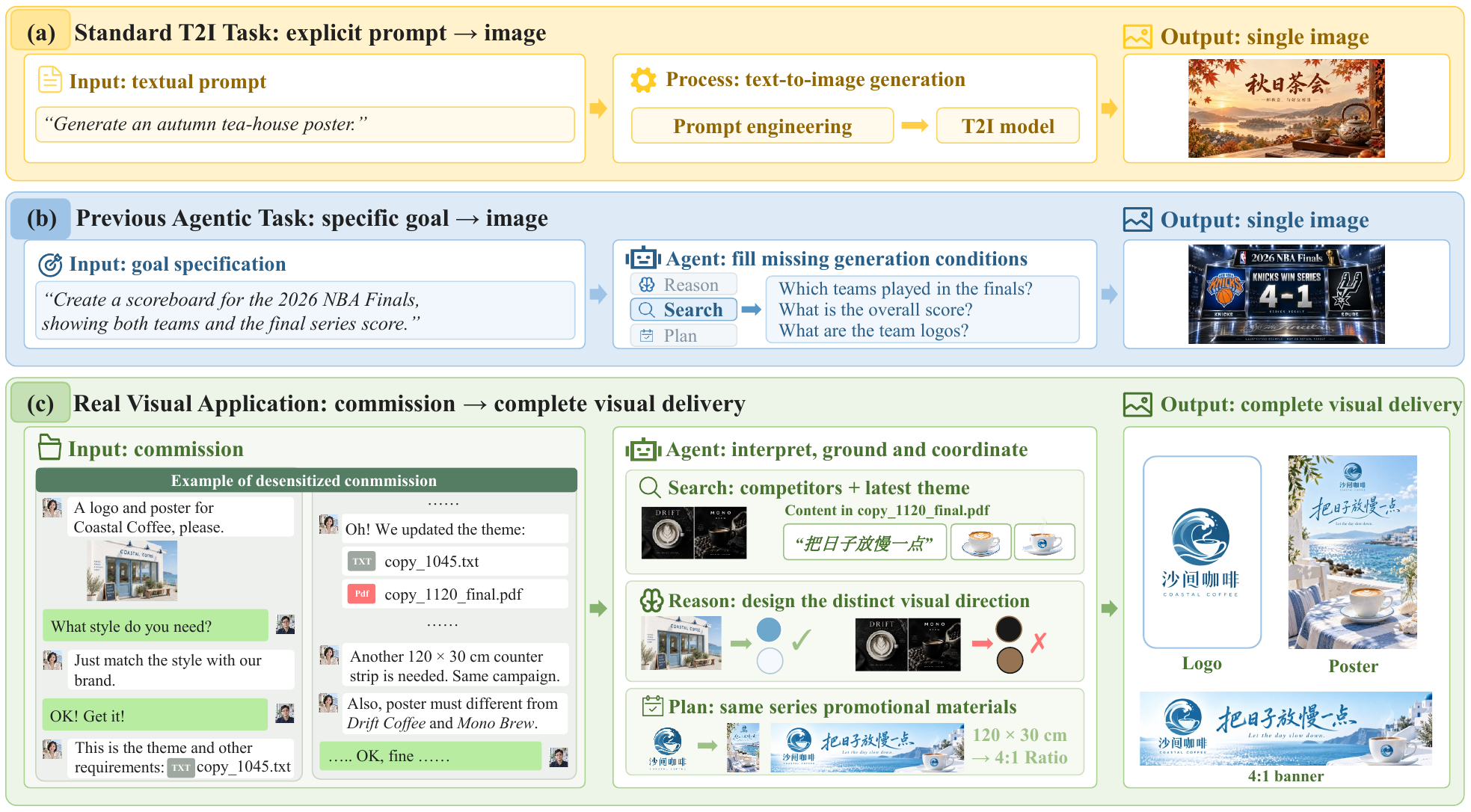}
\caption{\textbf{From image generation to visual commission completion.} (a). An explicit prompt guides image synthesis. (b). A specified generation goal is supplemented with missing information through reasoning, search, and planning. (c). A commission and its source materials require the agent to determine content and design, preserve active constraints, and coordinate the requested deliverables. VCB evaluates content correctness, visual realization, constraint preservation, and delivery completeness. The panels illustrate task settings rather than mutually exclusive method categories.
}
\label{fig:vwb_protocol}
\end{figure*}

Modern image models follow detailed instructions~\cite{wu2025qwen} to produce high-quality visuals, with growing use in visual production tasks. A visual commission, however, may begin with incomplete requirements and heterogeneous source materials and involve multiple dependent stages before final delivery. For example, preparing promotional materials for a tea house may require identifying valid content, deciding how to present it across different assets, preserving relevant constraints through generation and revision, and producing a complete and coherent set of deliverables. Much of this process is still coordinated manually, while text-to-image generation serves as only one part of the workflow. We study whether image agents can carry out such visual commissions, from interpreting source materials through to producing the required deliverables, within a given execution budget. 
Figure~\ref{fig:vwb_protocol} illustrates representative tasks and deliverables.


Existing evaluations have extended image generation beyond direct prompt following by introducing tasks that require reasoning, search, or planning~\cite{zhang2026qwen,ren2026scope,tang2026igenbench,ki2025graphicbench}. 
These settings provide valuable tests of capabilities and execution demands involved in visual production. 
A further question is whether success in these settings translates into reliable completion of visual commissions, where agents must integrate requirements and evidence distributed across source materials, carry intermediate decisions and constraints through dependent subtasks, and coordinate multiple deliverables. Evaluating these demands together allows us to examine how constituent capabilities support completion of the overall commission.

To address this question, we introduce \textbf{Visual Commission Bench (VCB)}, comprising 600 tasks across 12 categories and three complementary tiers: Specific-Capability (S), Multi-Stage (M), and Commission-Level (C). 
S tasks isolate capabilities relevant to visual task completion, such as reasoning, planning, information extraction, and retrieval. 
M tasks examine how intermediate results, decisions, and constraints are carried across dependent stages. 
C tasks evaluate complete visual commissions from source materials to coordinated delivery. 
The three tiers distinguish evaluation scope and task dependencies rather than intrinsic difficulty or tool-call counts, allowing us to study both individual capabilities and how they contribute to end-to-end task completion. All tasks are evaluated through their delivered artifacts without prescribing an internal workflow. 
5,039 carefully designed requirement checks are used to measure requirement-level fulfillment and derive task-level success.

VCB further reveals execution challenges that become salient at the commission level.
Some trajectories make little observable task progress despite substantial interaction, with agents repeatedly revisiting guidance and source materials, checking unchanged states, or recovering from similar interface errors. 
These behaviors suggest that commission-level execution depends not only on the capabilities available to the agent, but also on how appropriate task procedures are selected, reused, and coordinated throughout execution~\cite{li2026skillsbench,liu2026well,dong2026agent,tan2026skill}. This motivates us to examine structured support for both task guidance and execution management.
Preliminary runs on VCB reveal execution difficulties in carrying visual commissions through to completion.

We present \textbf{V-Protocol} as a reference agent for end-to-end visual commission completion. V-Protocol organizes reusable task knowledge into task, cognitive, and expression modules, and dynamically selects compatible combinations based on the request.
Task modules specify deliverables and execution dependencies, cognitive modules derive content from the inputs, and expression modules produce layout and rendering specifications. Selected modules exchange structured intermediate artifacts, while a shared harness~\cite{pan2026natural,dantanarayana2026sigil,lu2026contractskill,xia2026grasp} manages stage transitions, tool execution, and completion checks. Together, these components support controlled study of visual commission completion while keeping the underlying foundation models fixed.

\section{Related Work}
\label{sec:related_work}

\paragraph{Benchmarks for image agents.} WISE evaluates world knowledge integration~\cite{niu2026wise}, while Mind-Bench and KnowGen assess reasoning and search in image generation~\cite{he2026mind,feng2026gen}. Gen-Arena measures entity and constraint satisfaction~\cite{ren2026scope}, and IA-Bench covers planning, reasoning, search, and memory~\cite{zhang2026qwen}. GraphicBench and IGenBench examine graphic design and infographic reliability, respectively~\cite{ki2025graphicbench,tang2026igenbench}. VCB evaluates complete visual commissions that require agents to integrate distributed source materials, preserve requirements across dependent stages, and coordinate multiple deliverables. It assesses the delivered artifacts without prescribing an internal workflow.

\paragraph{Agentic image generation.} Gen-Searcher and Mind-Brush integrate retrieval and reasoning into generation~\cite{feng2026gen,he2026mind}. GenEvolve learns tool use from visual experience~\cite{chen2026genevolve}, and GEMS integrates memory and on-demand skills~\cite{he2026gems}. SCOPE maintains evolving semantic specifications for conditional skill orchestration~\cite{ren2026scope}, while Qwen-Image-Agent constructs generation context through planning and grounding~\cite{zhang2026qwen}. Other approaches use executable sketches~\cite{ye2026genclaw}, adaptive workflow routing~\cite{chen2026genrouter}, or evolving skill harnesses~\cite{li2026comfyclaw}. V-Protocol coordinates reusable task, cognitive, and expression modules through structured intermediate artifacts, linking content derived from source materials to its use across visual deliverables.

\paragraph{Skills and execution management.} AgentSkillOS and SkillSelect-Serve address skill organization and budget-aware selection~\cite{li2026organizing,zheng2026skillselect}; SkillSmith models complementarity and conflict among skills~\cite{wei2026skillsmith}. GraSP composes skills into typed dependency graphs with verification and local repair~\cite{xia2026grasp}. Natural-language harnesses expose reusable execution policies~\cite{pan2026natural}, while SIGIL and ContractSkill make procedural guidance executable and checkable~\cite{dantanarayana2026sigil,lu2026contractskill}. V-Protocol builds on these mechanisms to track artifact dependencies and completion conditions, leaving content and design decisions to the model. Visual checks assess requirement satisfaction alongside the harness's checks on execution state.
\section{Visual Commission Benchmark}
\label{sec:benchmark}


\textbf{Visual Commission Benchmark (VCB)} evaluates the
capabilities required for visual commission completion at different levels.
Tasks require agents to identify valid content from source materials,
decide how to present it across different assets, preserve
relevant constraints through generation and revision, and
produce a complete and coherent set of deliverables.
Evaluation is based on the submitted artifacts, covering content
correctness, visual realization, constraint preservation, and
delivery completeness without prescribing an internal workflow.
\Cref{tab:benchmark_comparison} summarizes how VCB relates to
existing evaluation settings. Figure~\ref{fig:v_protocol} in the appendix provides an overview of V-Protocol, including adaptive module composition, protocol execution, and artifact revision management.

\begin{table*}[t]
\centering
\caption{\textbf{Comparison of benchmark scale and evaluation focus.}
Checks denote task-specific verification items, while rating dimensions
denote graded evaluation criteria.
VCB broadens the evaluation toward end-to-end visual commission completion.
}
\label{tab:benchmark_comparison}
\begingroup
\small
\setlength{\tabcolsep}{6pt}
\renewcommand{\arraystretch}{1.15}
\begin{tabularx}{\linewidth}{@{}l r l X@{}}
\toprule
Benchmark
& \# Tasks
& Checks / Ratings
& Primary focus \\
\midrule
WISE
& 1,000
& 3-dim ratings
& World-knowledge reasoning \\

Mind-Bench
& 500
& 1,633 checks
& Reasoning and search \\

KnowGen
& 630
& 4-dim ratings
& Search-grounded generation \\

Gen-Arena
& 300
& 2,533 checks
& Entity and constraint satisfaction \\

IA-Bench$^{1}$
& 730
& 1,801 checks
& Planning, reasoning, search, and memory \\

\midrule
\textbf{VCB (Ours)}
& \textbf{600}
& \begin{tabular}[t]{@{}l@{}}
    \textbf{5,039 checks}\\
    \textbf{+ 4-dim ratings}
  \end{tabular}
 & \begin{tabular}[t]{@{}l@{}}\textbf{Distributed-source integration, dependent subtask}\\\textbf{execution, and coordinated multi-artifact delivery}\end{tabular}
 \\
\bottomrule
\end{tabularx}

\vspace{3pt}
\begin{minipage}{\linewidth}
\footnotesize
$^{1}$ IA-Bench is a closed source internal evaluation benchmark.
\end{minipage}
\endgroup
\end{table*}

\subsection{Task Scope and Public Inputs}
\label{sec:benchmark_tasks}

\begin{figure*}[t]
\centering
\includegraphics[width=\linewidth]{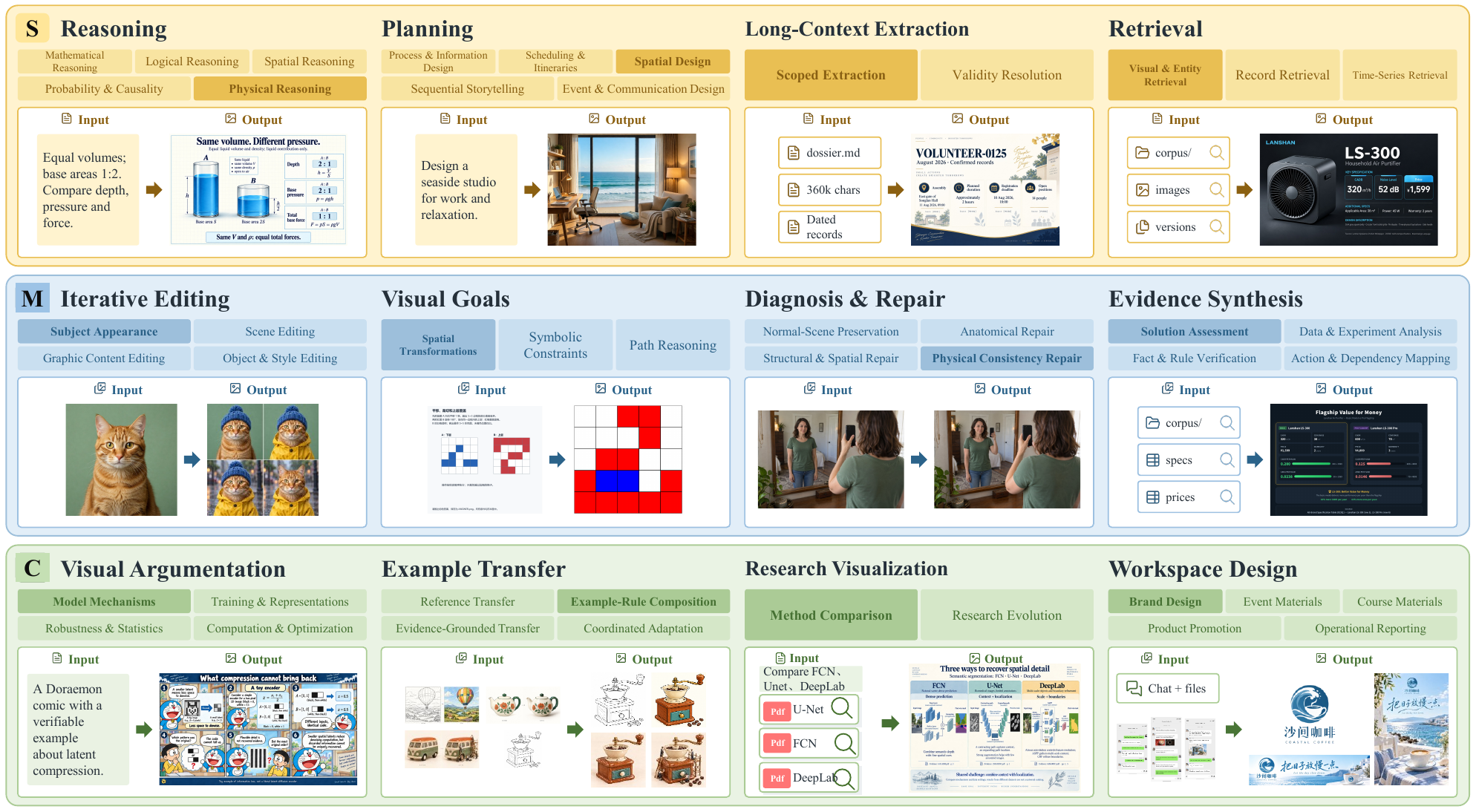}
\caption{\textbf{Overview of Visual Commission Benchmark (VCB).}
VCB contains 600 tasks across three tiers, 12 categories, and 45 subcategories.
The S tier focuses on reasoning, planning, long-context extraction, and retrieval.
The M tier combines capabilities through iterative editing, visual goal interpretation, diagnosis and repair, and evidence synthesis.
The C tier evaluates integrated visual commissions through visual argumentation, example transfer, research visualization, and workspace design.
Each panel presents an example input and output, with darker labels indicating the represented subcategories.
Documents and workspaces are shown schematically, and the displayed outputs are illustrative rather than experimental results.
}
\label{fig:vwb_bench}
\end{figure*}

As shown in Figure~\ref{fig:vwb_bench}, VCB comprises 600 tasks across 12 categories, with 50 tasks per category, organized into three complementary tiers: Specific-Capability (S), Multi-Stage (M), and Commission-Level (C). 
S tasks emphasize a focal capability: reasoning, planning, long-context information extraction, or retrieval.
M tasks examine how capabilities work together through iterative editing, visual goal interpretation, diagnosis and repair, and evidence synthesis.
C tasks involve visual argumentation, example-driven transfer, scientific visualization, and workspace-based design, requiring agents to make content and presentation decisions and organize the work needed for a complete delivery.
The tiers provide a progression in evaluation scope and capability coordination without prescribing a fixed workflow or number of model and tool calls.

Each task provides a user request, access to its source materials, and designated output locations.
Materials may include documents, tables, reference images, demonstrations, interaction histories, and workspace files.
Task-relevant information and constraints may be distributed across the source materials, requiring agents to identify and integrate them.
A task may require a single image, a sequence of edited images, or a coordinated set of visual artifacts.
Private evaluation annotations specify how to assess requirements grounded in the request and materials, without adding undisclosed task goals or prescribing a unique valid solution.
All tasks are self-contained evaluation packages and do not require additional user clarification during execution.

\subsection{Construction and Annotation}
\label{sec:benchmark_construction}

\paragraph{Task sources and diversity.}
VCB draws on hard cases from existing text-to-image~\cite{QIB} and
image-agent benchmarks~\cite{niu2026wise,he2026mind,feng2026gen}, newly constructed tasks, everyday user requests, and completed design commissions.
Real-world cases are anonymized and adapted into evaluation tasks while preserving their information dependencies, requirement changes, and delivery constraints.
Diversity spans both content domains and task structures, including information selection, dependent operations, constraint preservation, and multi-artifact coordination.
The resulting benchmark contains 600 tasks across 12 taskcategories and 45 subcategories, with 50 tasks per category and an equal split of 300 Chinese and 300 English tasks.


\paragraph{Task construction.}
Source cases are organized into evaluation tasks comprising a user request, source materials, and delivery requirements.
Relevant information may be distributed across documents, images, demonstrations, and interaction histories, requiring agents to identify and integrate it.
In total, VCB provides 4,288 source files containing approximately 9.81 million tokens of machine-readable text across task-specific materials and shared corpora.
Token counts use \texttt{cl100k\_base} and exclude text embedded in images.
Existing design outcomes can inform task construction and review.
For tasks that permit alternative designs, they serve as references rather than the only acceptable solutions.

\paragraph{Checklist construction and review.}
For each task, multiple models independently construct candidate checklists from its request and source materials. 
The participating models include GPT5.6-Sol, GPT6-Astra, GLM-5.3-Flash, and DeepSeek-v4.1-Flash.
The candidate checklists are consolidated by merging overlapping checks, retaining complementary requirements, and resolving conflicting criteria.
Each resulting check expresses a concrete judgment question and specifies the supporting evidence and images needed for evaluation.
The models then assess the consolidated checks individually for correctness, grounding, and clarity, and examine the checklists for unsupported requirements, redundancy, and omissions.
Human reviewers examine the proposed checks and model judgments to identify necessary revisions.
The checklists are refined through repeated rounds of model assessment, human review, and revision.

\subsection{External Evaluation}
\label{sec:benchmark_evaluation}

\paragraph{ReqPass and TaskPass.}
Evaluation is based on the artifacts saved at the designated
output locations.
Let $\mathcal{Q}_n$ denote the applicable checklist items
for task $n$, and let $r_{nj}\in\{0,1\}$ indicate whether
item $j$ passes.
Let $v_n\in\{0,1\}$ denote the task-level delivery check,
which verifies that all required artifacts are saved as
specified and satisfy the applicable file-level requirements.
We report requirement pass rate and strict task pass rate:
\begin{equation}
\begin{aligned}
\mathrm{ReqPass}
&=
100\,
\frac{
\sum_{n=1}^{N}
\left(v_n+\sum_{j\in\mathcal{Q}_n}r_{nj}\right)
}{
N+\sum_{n=1}^{N}|\mathcal{Q}_n|
},\\
\mathrm{TaskPass}
&=
\frac{100}{N}
\sum_{n=1}^{N}
v_n\prod_{j\in\mathcal{Q}_n}r_{nj}.
\end{aligned}
\label{eq:benchmark_metrics}
\end{equation}
ReqPass pools applicable checklist items and one delivery
check per task, while TaskPass requires all of them to pass.
Checks that depend on a missing or invalid artifact fail;
checks on other available artifacts remain eligible for
partial credit.
Conditional checks are included only when applicable,
and revision checks follow the requirements active at
the corresponding checkpoint.

\paragraph{Visual quality.}
Visual quality (VQ) assesses readability, layout,
visual coherence, and aesthetic appeal separately
from factual correctness.
Each assessable output image receives a rating
from 1 to 5 on each dimension.
Let $\mathcal{O}_n$ be the set of required output images
for task $n$, and let $s_{nkd}$ be the rating of image $k$
on dimension $d$.
We report
\begin{equation}
\mathrm{VQ}
=
\frac{1}{N}
\sum_{n=1}^{N}
\frac{1}{|\mathcal{O}_n|}
\sum_{k\in\mathcal{O}_n}
\frac{1}{4}\sum_{d=1}^{4}s_{nkd}.
\label{eq:benchmark_vq}
\end{equation}
Missing or undecodable required images receive zero
on all four dimensions.
Ratings are averaged within each task before being
averaged across tasks, giving each task equal weight.

\paragraph{Judgment protocol.}
File-level properties, such as existence, decodability,
and specified dimensions, are checked programmatically.
Qwen3.8-Flash-Next serves as the evaluator for both
requirement satisfaction and visual quality, with separate
inputs and criteria.
For requirement evaluation, it receives the checklist
question, its supporting criteria, and the images needed
for that judgment.
Source facts are incorporated into the criteria, while
reference images are supplied when visual comparison
is necessary.
For VQ evaluation, it receives only the output image
and the fixed quality rubric.
Both evaluations are blinded to the identity of the
evaluated system and exclude its execution trace.
Evaluator failures or unresolved judgments are recorded
separately and retried rather than treated as task failures
or assigned zero VQ.
Final scores are reported once all required judgments
have been resolved.
\section{V-Protocol}
\label{sec:method}
V-Protocol composes reusable modules into executable programs for visual commissions. The model selects procedures and determines content and design; a shared harness manages artifact dependencies, tool execution, and completion checks.

\subsection{Module Representation}
\label{sec:module_representation}
Task modules specify commission procedures and optional program templates; cognitive modules guide material interpretation, retrieval, reasoning, and synthesis; expression modules provide layout, style, and rendering guidance. Each module declares its family, capabilities, input and output types, applicability conditions, and procedural guidance. Registry metadata supports selection, with full guidance loaded on activation. Declared interfaces identify possible handoffs, while the model assesses their suitability for the current task.

\subsection{Adaptive Module Composition}
\label{sec:module_composition}
The agent derives a task specification from the public request and inspected materials, recording the goal, inputs, active requirements, and deliverables. A suitable task template initializes the program; otherwise, the model constructs it directly. At step $t$, the program is
\begin{equation}
\mathcal{P}_t=(G_t,\mu_t), \qquad G_t=(\mathcal{V}_t,\mathcal{E}_t),
\end{equation}
where nodes represent operations, edges record declared artifact dependencies, and $\mu_t$ records module assignments. Each operation specifies its objective, input and output artifacts, capability needs, and permitted actions. Source materials can supply inputs directly. Operations may use several modules, and modules may be reused across operations.

For each operation, the registry retrieves modules whose capabilities overlap its needs and orders candidates by configured priority and capability coverage. The model selects applicable guidance for the current objective and inputs, revising the program as inspection and generation reveal additional needs. Simple requests can use a minimal program. M

\subsection{Protocol Execution}
\label{sec:protocol_execution}
The harness validates the operation graph and artifact references. Before each action, it checks input availability, dependency freshness, permissions, and remaining budget. The model supplies content and tool arguments; the harness mediates calls and registers outputs. Node completion requires recorded artifacts and valid tool results. Program updates retain execution history and valid completed outputs while allowing future operations to change. Retries and repairs consume the same task budget.

Artifacts carry identifiers, revisions, and declared source dependencies. Dependency snapshots record the source versions and content hashes used by downstream operations. A change to a declared source makes the corresponding snapshot stale, preventing affected artifacts from being reused as current evidence until the dependency is resolved. Image edits record their source revision, and replacing an image at the same path creates a new revision.

Online verification compares the current image with active requirements and content checkpoints. Reports bind to the image identity, revision, content hash, and verification criteria; changes to the image or relevant criteria invalidate prior reports. The model uses feedback to choose edits, regeneration, or program revisions. Before delivery, the harness checks that required artifacts exist at their designated locations and satisfy runtime completion conditions. These checks concern recorded requirements and dependencies; semantic judgments remain the model's responsibility. Private benchmark annotations are unavailable during planning, verification, and repair, and final artifacts are assessed through the external evaluation protocol.
\section{Experiments}
\label{sec:experiments}

\subsection{Experimental Setup}
\label{sec:experimental_setup}

We evaluate V-Protocol on VCB, WISE-Verified, and Mind-Bench.
We follow each benchmark's evaluation protocol.
For VCB, we report \textsc{ReqPass}, \textsc{TaskPass},
and visual quality (VQ), both overall and across
the S, M, and C tiers.
WISE-Verified uses binary WiScore, and Mind-Bench uses
checklist-based strict accuracy (CSA).

We compare against direct Qwen-Image2.1 generation,
Mind-Brush, GEMS, and SCOPE.
Agent-based methods share a Qwen3.8-27B controller
with medium thinking and a Qwen-Image2.1 image backend.
Where web search is permitted, they use DuckDuckGo.
Each method retains its own planning, generation,
and revision procedure.
Qwen3.8-Flash-Next evaluates VCB requirement satisfaction
and visual quality using separate prompts.
Published reference results are presented separately
from experiments with shared backends.
Additional adaptation, evaluation, and recovery details
are provided in Appendix~\ref{app:exp_protocol}.

\subsection{Results and Analysis}
\label{sec:main_results}

\begin{table*}[t]
\centering
\caption{\textbf{Commission completion on VCB.}
Each tier reports \textsc{ReqPass} (RP),
\textsc{TaskPass} (TP), and visual quality (VQ).
RP and TP are percentages.
VQ averages 1--5 ratings, with missing or undecodable
required images assigned zero.
Overall reports benchmark-wide results.
V-Protocol achieves the highest scores, while complete
commissions remain challenging even when visual quality
is high.}
\label{tab:vcb_main}
\begingroup
\small
\setlength{\tabcolsep}{3pt}
\renewcommand{\arraystretch}{1.12}
\resizebox{\linewidth}{!}{%
\begin{tabular}{@{}l*{12}{r}@{}}
\toprule
& \multicolumn{3}{c}{S}
& \multicolumn{3}{c}{M}
& \multicolumn{3}{c}{C}
& \multicolumn{3}{c}{Overall} \\
\cmidrule(lr){2-4}
\cmidrule(lr){5-7}
\cmidrule(lr){8-10}
\cmidrule(lr){11-13}
System
& RP$\uparrow$ & TP$\uparrow$ & VQ$\uparrow$
& RP$\uparrow$ & TP$\uparrow$ & VQ$\uparrow$
& RP$\uparrow$ & TP$\uparrow$ & VQ$\uparrow$
& RP$\uparrow$ & TP$\uparrow$ & VQ$\uparrow$ \\
\midrule
Direct
& 8.33 & 0.50 & 2.396
& 12.15 & 0.00 & 2.039
& 12.81 & 0.00 & 2.091
& 11.14 & 0.17 & 2.175 \\
Mind-Brush
& 13.83 & 1.00 & 2.794
& 6.71 & 0.00 & 2.010
& 15.32 & 0.00 & 2.121
& 12.69 & 0.33 & 2.308 \\
GEMS
& 41.20 & 16.00 & 3.639
& 10.49 & 0.00 & 2.236
& 34.69 & 8.00 & 2.337
& 30.92 & 8.00 & 2.737 \\
SCOPE
& 84.36 & 55.00 & 3.658
& 77.74 & 39.50 & 4.097
& 82.10 & 31.50 & 4.192
& 81.78 & 42.00 & 3.982 \\
\midrule
\textbf{V-Protocol}
& \textbf{96.95} & \textbf{82.00} & \textbf{4.250}
& \textbf{91.78} & \textbf{67.00} & \textbf{4.157}
& \textbf{94.15} & \textbf{58.00} & \textbf{4.195}
& \textbf{94.48} & \textbf{69.00} & \textbf{4.200} \\
\bottomrule
\end{tabular}%
}
\endgroup
\end{table*}

\paragraph{VCB.}
Table~\ref{tab:vcb_main} shows that V-Protocol achieves
the highest overall scores, with 94.48\% \textsc{ReqPass},
69.00\% \textsc{TaskPass}, and 4.200 VQ.
Compared with SCOPE, the strongest baseline overall,
V-Protocol improves \textsc{ReqPass} by 12.70 percentage
points and \textsc{TaskPass} by 27.00 percentage points.
Direct generation, Mind-Brush, and GEMS complete
substantially fewer tasks.

\begin{figure*}[t]
\centering
\includegraphics[width=\linewidth]
{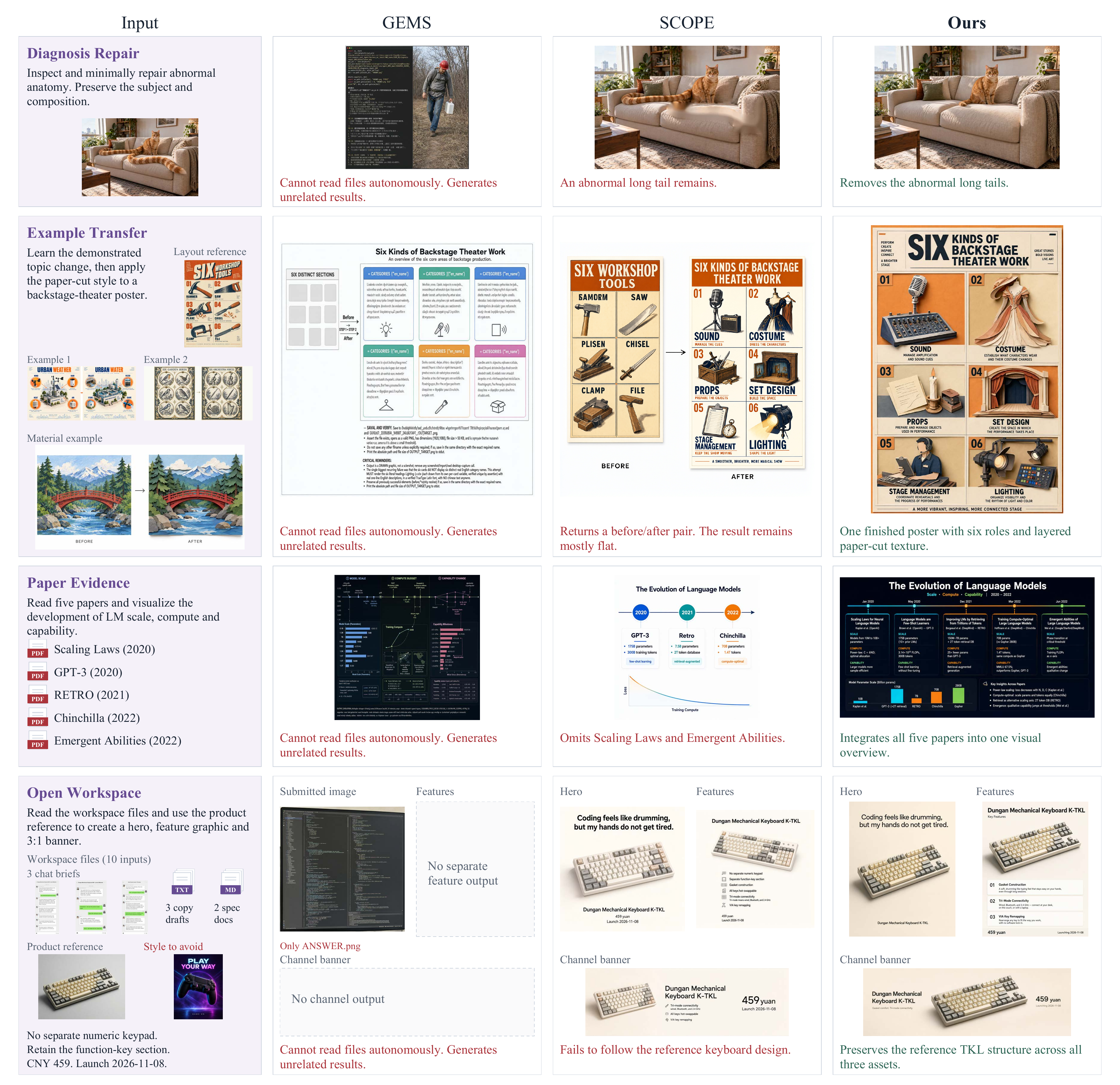}
\caption{\textbf{Qualitative comparison.}
V-Protocol achives better visual commissions completion.
}
\label{fig:vwb_compare}
\end{figure*}

Completion remains challenging as task scope expands.
V-Protocol's \textsc{TaskPass} decreases from 82.00\%
on S to 67.00\% on M and 58.00\% on C.
SCOPE follows the same trend.
V-Protocol's C-tier \textsc{ReqPass} of 94.15\%
also contrasts with its 58.00\% \textsc{TaskPass},
highlighting the remaining challenge of satisfying
all requirements within a commission. Figure~\ref{fig:vwb_compare} complements the quantitative
results with a qualitative comparison of visual outputs
on selected VCB tasks.

\begin{table*}[t]
\centering
\caption{\textbf{Results on WISE-Verified and Mind-Bench.}
V-Protocol achieves the highest overall scores among
the compared systems, with a larger advantage over SCOPE
on reasoning-driven Mind-Bench tasks.
WISE-Verified reports binary WiScore, and Mind-Bench
reports CSA.
Knowledge and Reasoning denote the corresponding
category averages.
The first three rows report published references;
the remaining rows use shared backends.
Bold marks the best results within the shared-backend group.}
\label{tab:public_main}
\begingroup
\small
\setlength{\tabcolsep}{3pt}
\renewcommand{\arraystretch}{1.12}
\resizebox{\linewidth}{!}{%
\begin{tabular}{@{}l*{10}{r}@{}}
\toprule
& \multicolumn{7}{c}{WISE-Verified}
& \multicolumn{3}{c}{Mind-Bench} \\
\cmidrule(lr){2-8}
\cmidrule(lr){9-11}
System
& Culture & Time & Space & Biology & Physics & Chemistry
& Overall
& Knowledge & Reasoning & Overall \\
\midrule
GPT-Image-1.5
& 0.8900 & 0.6917 & 0.8833 & 0.8000 & 0.7583 & 0.7750
& 0.8250
& 0.220 & 0.180 & 0.210 \\
Nano Banana Pro
& 0.8975 & 0.8167 & 0.9333 & 0.8167 & 0.8667 & 0.8750
& 0.8760
& 0.400 & 0.440 & 0.410 \\
Qwen-Image-Agent
& 0.9200 & 0.9167 & 0.9333 & 0.8333 & 0.8667 & 0.9000
& 0.9020
& 0.404 & 0.436 & 0.420 \\
\midrule
Direct
& 0.4800 & 0.4667 & 0.5583 & 0.3417 & 0.5417 & 0.3083
& 0.4580
& 0.032 & 0.052 & 0.042 \\
Mind-Brush
& 0.6550 & 0.6417 & 0.6750 & 0.3750 & 0.5583 & 0.3500
& 0.5740
& 0.136 & 0.060 & 0.098 \\
GEMS
& 0.8625 & 0.7833 & 0.9000 & 0.7667 & 0.7917 & 0.7167
& 0.8200
& 0.228 & 0.336 & 0.282 \\
SCOPE
& \textbf{0.9175} & \textbf{0.9333} & \textbf{0.9083}
& 0.8583 & 0.8500 & 0.9000
& 0.9010
& \textbf{0.368} & 0.416 & 0.392 \\
\midrule
\textbf{V-Protocol}
& 0.9100 & 0.9083 & 0.9000
& \textbf{0.8750} & \textbf{0.9417} & \textbf{0.9167}
& \textbf{0.9090}
& \textbf{0.368} & \textbf{0.540} & \textbf{0.454} \\
\bottomrule
\end{tabular}%
}
\endgroup
\end{table*}

\paragraph{Public benchmark for image agents.}
Table~\ref{tab:public_main} shows that V-Protocol also
achieves the highest overall scores among the compared
systems on WISE-Verified and Mind-Bench.
WISE-Verified has a more focused task scope than VCB,
and leading systems already perform well.
V-Protocol scores 0.9090, narrowly ahead of SCOPE's 0.9010.

The difference is larger on Mind-Bench, where V-Protocol
achieves 0.454 CSA compared with 0.392 for SCOPE.
Their knowledge-driven averages are identical at 0.368,
while V-Protocol improves the reasoning-driven average
from 0.416 to 0.540.
Its advantage on Mind-Bench is therefore concentrated
in tasks requiring reasoning to determine the visual content.
Together with the larger completion gap on VCB,
these results show that comparable performance on
knowledge-oriented generation does not necessarily imply
comparable ability to complete visual commissions.

\begin{table*}[t]
\centering
\begin{minipage}[t]{0.57\textwidth}
\vspace{0pt}
\textbf{Component analysis.}
We examine runtime enforcement and adaptive module selection
on VCB.
The runtime variant disables execution gates while
retaining module selection and tool access.
The selection variant replaces task-adaptive module
choices with a predefined module set.

\medskip
As shown in Table~\ref{tab:ab_main}, the two variants
achieve 60.17\% and 60.83\% \textsc{TaskPass},
respectively, compared with 69.00\% for the Full reference.
Both also obtain lower \textsc{ReqPass} and VQ.
These comparisons are exploratory because Full reuses
historical outputs, including recovery runs, whereas
the variants use a later frozen runtime.
Detailed controls and additional results are provided
in Appendix~\ref{app:ablation_protocol}.
\end{minipage}
\hfill
\begin{minipage}[t]{0.40\textwidth}
\vspace{0pt}
\centering
\caption{\textbf{Component analysis on VCB.}}
\label{tab:ab_main}
\begingroup
\small
\setlength{\tabcolsep}{3pt}
\renewcommand{\arraystretch}{1.18}
\resizebox{\linewidth}{!}{%
\begin{tabular}{@{}lrrr@{}}
\toprule
Configuration
& RP$\uparrow$ & TP$\uparrow$ & VQ$\uparrow$ \\
\midrule
\shortstack[l]{w/o Runtime\\Enforcement}
& 82.14 & 60.17 & 3.800 \\ 
\addlinespace[3pt]
\shortstack[l]{w/o Adaptive\\Module Selection}
& 82.20 & 60.83 & 3.862 \\ 
\midrule
Full V-Protocol
& 94.48 & 69.00 & 4.200 \\
\bottomrule
\end{tabular}%
}
\endgroup
\end{minipage}
\end{table*}
\section{Conclusion}
\label{sec:conclusion}
We introduced Visual Commission Bench (VCB) to evaluate image agents on complete visual commissions, from interpreting distributed source materials to delivering coordinated artifacts. V-Protocol supports these tasks by composing reusable task, cognitive, and expression modules through structured intermediate artifacts, while an execution harness tracks dependencies, revisions, and completion. With shared model backbones, V-Protocol achieves higher task success than the evaluated baselines. The commission-level results reveal substantial differences in completion despite comparable visual quality, demonstrating the importance of evaluating fulfillment of the entire request. Complete commissions remain challenging even with explicit execution support. Reliable use of intermediate results and preservation of requirements across revisions and deliverables remain priorities for further work.

\bibliography{preprint}
\bibliographystyle{preprint}

\clearpage
\appendix
\begin{center}
{\LARGE\bfseries Supplementary Material}
\end{center}

\section{V-Protocol}
\label{sec:app_method}
\begin{figure}[t]
\centering
\includegraphics[width=\linewidth]{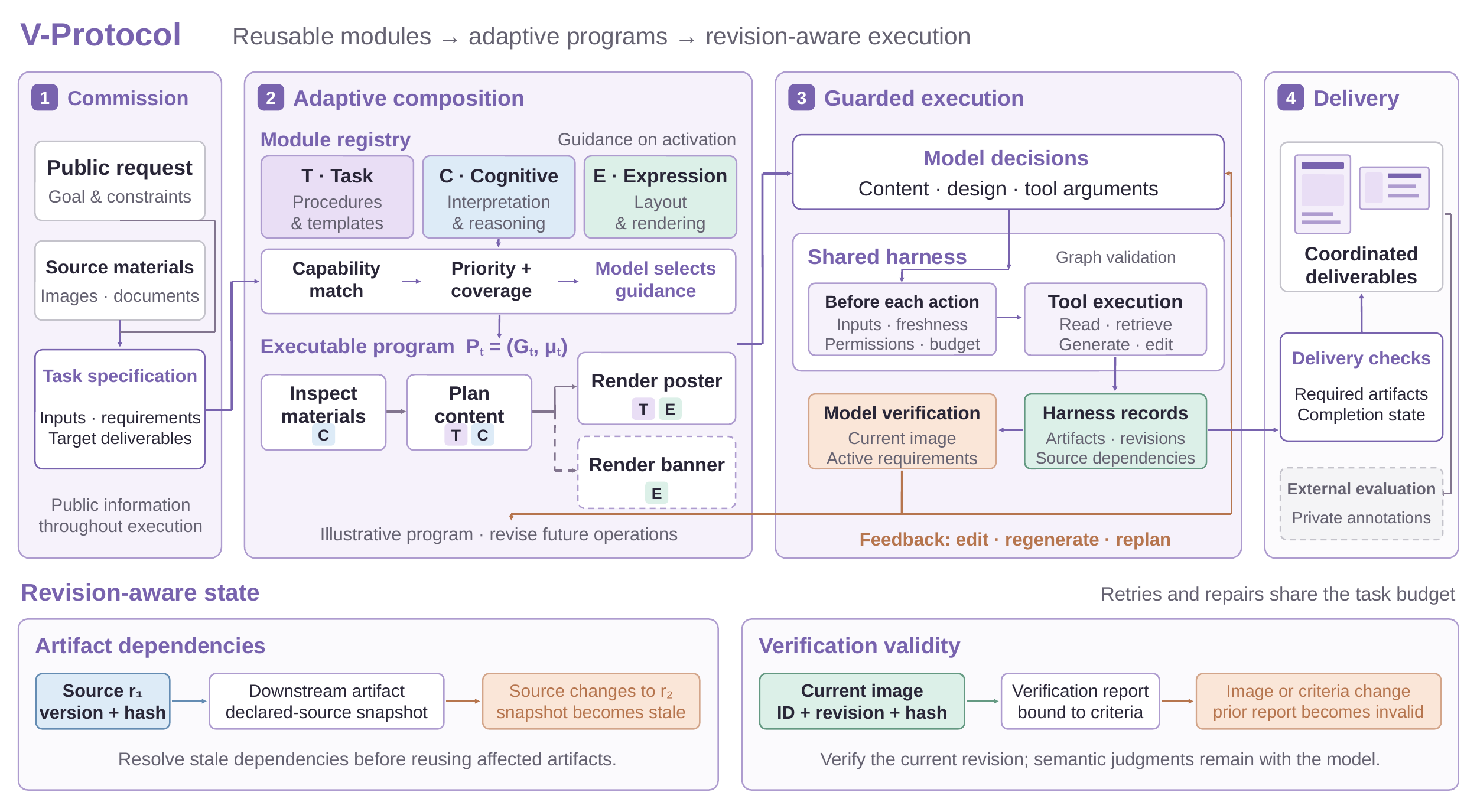}
\caption{\textbf{Overview of V-Protocol.} Public requests and source materials define a task specification, from which the agent composes an executable program using task (T), cognitive (C), and expression (E) modules. Colored tags indicate module assignments; the dashed branch illustrates a revisable future operation. The model determines content, design, and tool arguments, while the harness checks runtime conditions and records artifacts. Verification feedback informs edits, regeneration, or program revision. The bottom panels illustrate how changes to declared sources make dependency snapshots stale, while changes to an image or relevant criteria invalidate its verification reports. Private benchmark annotations are used only for external evaluation.}
\label{fig:v_protocol}
\end{figure}

V-Protocol composes reusable modules into executable programs for visual commissions (Figure~\ref{fig:v_protocol}). The model selects applicable guidance, determines content and design, and revises future operations using inspected materials and verification feedback. A shared harness mediates tool execution, tracks declared artifact dependencies and revisions, and checks runtime completion conditions. Semantic judgments remain with the model.

\section{Experimental Setup and Evaluation Metrics}
\label{app:exp_protocol}

\subsection{Benchmarks and Evaluation Metrics}
\label{app:benchmark_tracks}

VCB contains 600 tasks across three tiers, with 200 tasks per tier
and 50 tasks in each of the 12 categories.
\textsc{ReqPass} pools applicable checklist items and one delivery
check per task. \textsc{TaskPass} measures the fraction of tasks
satisfying all applicable checks and the delivery check.
VQ averages four ratings within each required image, then averages
images within a task and tasks within the benchmark.
Thus, \textsc{TaskPass} and VQ weight tasks equally, whereas
\textsc{ReqPass} weights applicable checks equally.
Overall \textsc{ReqPass} is computed from pooled counts rather than
the arithmetic mean of the tier scores.

WISE-Verified contains 1,000 requests across Culture, Time, Space,
Biology, Physics, and Chemistry. Its overall binary WiScore weights
Culture by 0.40 and each remaining category by 0.12.
Mind-Bench contains 500 requests across ten equally sized categories.
A task passes checklist-based strict accuracy (CSA) only when all
applicable checklist conditions pass.
Overall CSA is computed across the 500 requests, while the
knowledge-driven and reasoning-driven averages each cover five
categories. Results are reported separately for the three benchmarks.

\subsection{Backends and Execution Settings}

The agent controller is Qwen3.8-27B with medium thinking, and the
image backend is Qwen-Image2.1. Web-search calls use DuckDuckGo where
permitted. VCB retrieval uses the benchmark-provided local corpora.
Direct generation uses the image backend without an agent controller.
The agent methods retain their own procedures for planning, tool use,
generation, and revision. Shared backends do not imply equal numbers
of model calls, image generations, or execution time.
Published reference results retain their original configurations
and are distinguished from the shared-backend comparisons.
For dated requests, the target is the time specified by the benchmark.

\section{Component Analysis and Module Budgets}
\label{app:ablation_protocol}



\paragraph{Runtime enforcement.}
The runtime variant retains module selection,
generation and editing tools, artifact registration,
input validation, access restrictions, and resource limits.
It disables specified execution gates, including
node-action authorization, generation prerequisites,
dependency-freshness checks, node-completion enforcement,
and final-delivery enforcement.
The intervention therefore removes selected execution
constraints rather than the entire runtime.
The agent may declare completion, but task success is
determined independently from its delivered artifacts.

\paragraph{Adaptive module selection.}
The fixed-selection variant uses the same predefined
module set for every request:
\texttt{task-deliverable-bundle},
\texttt{cognitive-evidence-synthesis},
\texttt{expression-basic}, and
\texttt{expression-hybrid-composition}.

\paragraph{Execution and evaluation.}
The new variants use the same controller and image
backend, with limits of 256 steps, eight generation
calls, and eight editing calls per task.
Their source code and module definitions are frozen.
Each task receives one execution, which may include
generation and revision steps within these limits.
No output-driven recovery run or task replacement
is added afterward.
Evaluation-service failures are retried separately.
Missing or invalid deliverables remain subject to
the benchmark's failure and zero-VQ rules.
\paragraph{Module-budget setting.}
We examine restricted module budgets with
$K \in \{1,2,3\}$.
The budget counts distinct module identifiers loaded
throughout a task across the task, cognitive, and
expression axes, including default expression guidance.
Reusing a module does not consume another slot,
whereas loading a new module at a later stage does.
The budget is an upper bound, not a requirement to
load exactly $K$ modules, and no minimum allocation
is imposed on any axis.

When the budget leaves no room for an expression
module, the agent may use the base instructions and
tool schemas without satisfying the expression-module
prerequisite.
Other artifact and execution constraints remain active.
Table~\ref{tab:module_budget} compares these selected
budget configurations with the historical Full reference.

\begin{table}[t]
\centering
\caption{\textbf{Selected module-budget results.}
$K$ caps the number of distinct modules loaded throughout
a task across all three axes.
RP and TP are percentages; VQ is measured on a 0--5 scale.
}
\label{tab:module_budget}
\begingroup
\small
\setlength{\tabcolsep}{5pt}
\renewcommand{\arraystretch}{1.12}
\begin{tabular}{lrrr}
\toprule
Configuration
& RP$\uparrow$ & TP$\uparrow$ & VQ$\uparrow$ \\
\midrule
Adaptive selection, $K=1$ 
    & 82.35 & 60.17 & 3.769 \\
Adaptive selection, $K=2$ 
    & 84.24 & 59.83 & 3.853 \\
Adaptive selection, $K=3$ 
    & 84.83 & 59.83 & 3.872 \\
\midrule
Full V-Protocol (reference)
& 94.48 & 69.00 & 4.200 \\
\bottomrule
\end{tabular}
\endgroup
\end{table}

\section{Execution Observations from SCOPE and V-Protocol}
\label{app:scope_runtime_diagnostics}

We examine retained execution traces of SCOPE~\cite{ren2026scope} and
V-Protocol on VCB (600 tasks), WISE-Verified (1,000 tasks), and Mind-Bench (500 tasks).
The analysis documents cases in which repeated controller actions
do not directly inspect task state or advance a visual artifact.
It provides evidence for execution difficulties in visual tasks
rather than isolating algorithmic causes.

\paragraph{Coverage and deployment.}
Shell-event counts use one retained log per task and method, rather
than a complete history of all attempts. Earlier overwritten attempts
and later manual recovery are not included in these trace statistics.
The observations characterize the evaluated executions and their
controller--tool interfaces; they do not by themselves identify
limitations intrinsic to the original SCOPE algorithm.

\paragraph{Controller-issued waiting and repetition.}
Table~\ref{tab:scope_runtime_events} reports explicit shell waiting and
repeated shell commands, using one retained log per task for each method.
SCOPE's VCB logs contain 5,386 sleep-containing shell records across 361
tasks and 2,652 repeated command records. V-Protocol's corresponding logs
contain no explicit shell sleep and 31 repeated commands across 28 tasks.
Neither method is assumed to have zero backend waiting: the table captures
controller-issued shell operations, not waits within tools.
Both methods use the same sleep-detection pattern,
\verb|\bsleep\s+\d|, on recorded shell command strings.
This lexical rule does not capture every possible waiting construct.
Repeated commands are counted within each task after whitespace
normalization, starting with the second occurrence. In particular,
V-Protocol's WISE-Verified and Mind-Bench traces contain no shell calls; their zeros
describe the execution interface, not an absence of service latency.

\begin{table*}[t]
\centering
\caption{\textbf{Observed controller-issued shell operations.}
Each method has one retained log per task. Sleep counts records containing
an explicit shell \texttt{sleep}; repeated commands count occurrences beyond
the first identical command within a task after whitespace normalization.
These are execution-pattern statistics, not counts of all wasted work.
Tool-internal waits are excluded.}
\label{tab:scope_runtime_events}
\small
\setlength{\tabcolsep}{8pt}
\resizebox{\linewidth}{!}{%
\begin{tabular}{@{}llrrrr@{}}
\toprule
Benchmark & Method & Tasks & Tasks with sleep & Sleep records & Repeated commands \\
\midrule
VCB & SCOPE & 600 & 361 & 5,386 & 2,652 \\
    & V-Protocol & 600 & 0 & 0 & 31 \\
\midrule
WISE-Verified & SCOPE & 1,000 & 519 & 5,136 & 1,325 \\
     & V-Protocol & 1,000 & 0 & 0 & 0 \\
\midrule
Mind-Bench & SCOPE & 500 & 358 & 8,970 & 3,829 \\
          & V-Protocol & 500 & 0 & 0 & 0 \\
\bottomrule
\end{tabular}%
}
\end{table*}

Repeated polling can be useful, so repetition alone is not evidence of
waste. A clearer example is VCB task \texttt{0378\_M4}: the SCOPE controller
submits \texttt{sleep 1; echo ok} 80 times, with 78 successful executions
and two argument-parsing failures. This command neither checks task state
nor produces a deliverable. The retained session contains 344 tool requests
over 63.91 minutes. In \texttt{0120\_S3}, 212 of 453 tool requests contain
sleep, and two wait-and-check commands each recur 12 times. These examples
establish the existence of low-progress interaction, not its prevalence
across all repeated calls. The first session's duration is not attributed
entirely to the repeated command.

\paragraph{Runtime intervention.}
V-Protocol's retained VCB logs include 82 rejections tagged as
no-progress submissions and four consecutive-identical-call guard
rejections. These events show that progress checks intervene during
execution. They do not establish that every rejected action is
followed by successful recovery. Argument and format errors also
occur in both systems, but their different controller--tool interfaces
do not provide a common basis for comparing raw error counts.

\paragraph{Controller usage and session duration.}
Table~\ref{tab:scope_runtime_cost} summarizes recorded
controller usage and session duration on VCB.
The analysis uses tasks with recoverable controller usage
and session timestamps for both methods, paired by task
identifier without filtering by task success.
Tasks without complete paired records are excluded from
this analysis but remain in the benchmark evaluation.

Token counts accumulate input and output across calls
within the retained primary-controller session, including
repeated context and cached input.
Session duration measures the interval between the first
and last recorded events, including tool waits.
It does not include elapsed time across earlier attempts
or subsequent manual recovery.

\begin{table*}[t]
\centering
\caption{\textbf{Recorded controller usage on VCB.}
Token counts are per-task means in thousands, with cached
input included.
Time measures the retained session's event span.
V-Protocol records lower token volumes and shorter median
and P90 session durations in these historical runs.}
\label{tab:scope_runtime_cost}
\small
\setlength{\tabcolsep}{8pt}
\resizebox{\linewidth}{!}{%
\begin{tabular}{@{}lrrrrr@{}}
\toprule
Method
& Input (k)
& Output (k)
& Total (k)
& Median time (min)
& P90 time (min) \\
\midrule
SCOPE
& 4,895.7 & 54.7 & 4,950.4 & 18.45 & 57.01 \\
V-Protocol
& 594.7 & 22.1 & 616.8 & 11.01 & 18.88 \\
\bottomrule
\end{tabular}%
}
\end{table*}

Cached input accounts for 96.2\% of SCOPE's recorded input
tokens, while V-Protocol's retained logs do not preserve
the cache split.
Token volumes therefore do not directly measure differences
in uncached computation or monetary cost.
The accounting covers the primary controller and excludes
separate image-model computation, external evaluation,
and auxiliary model calls not recorded in its usage.
Because concurrency, service load, and recovery history
were not matched, session durations are descriptive
observations rather than a controlled latency comparison.

\paragraph{Implication.}
The concrete cases show that substantial controller interaction
can occur without directly advancing a visual artifact.
These observations motivate support for execution progress and
artifact completion alongside reasoning about image content.
They do not establish that the repeated shell operations alone
explain the observed resource differences.

\clearpage
\section{Qualitative Examples}
\label{app:qualitative}

\subsection{Task Inputs and Outputs}
\label{app:visual_task_examples}

The following figures illustrate inputs and visual outputs for the
twelve task categories. Long materials are summarized by excerpts
or file representations for presentation.
These illustrations explain the task settings rather than estimate
benchmark performance.

\begin{figure*}[t]
\centering
\includegraphics[width=\linewidth]
{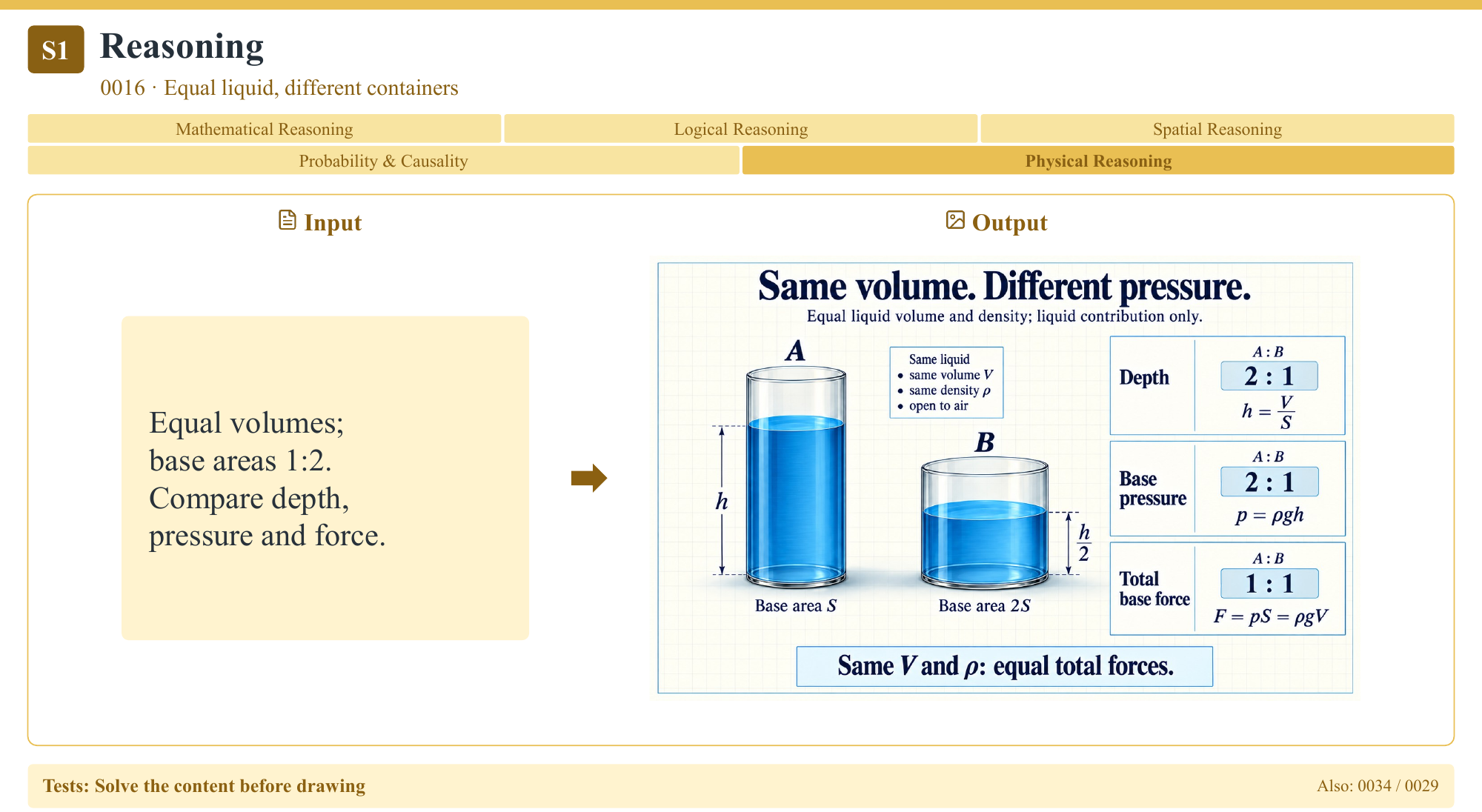}
\caption{\textbf{S1: Reasoning.} The example compares liquid depth, base pressure, and total force for equal liquid volumes in containers with different base areas.}
\label{fig:sp_task_s1}
\end{figure*}

\begin{figure*}[t]
\centering
\includegraphics[width=\linewidth]
{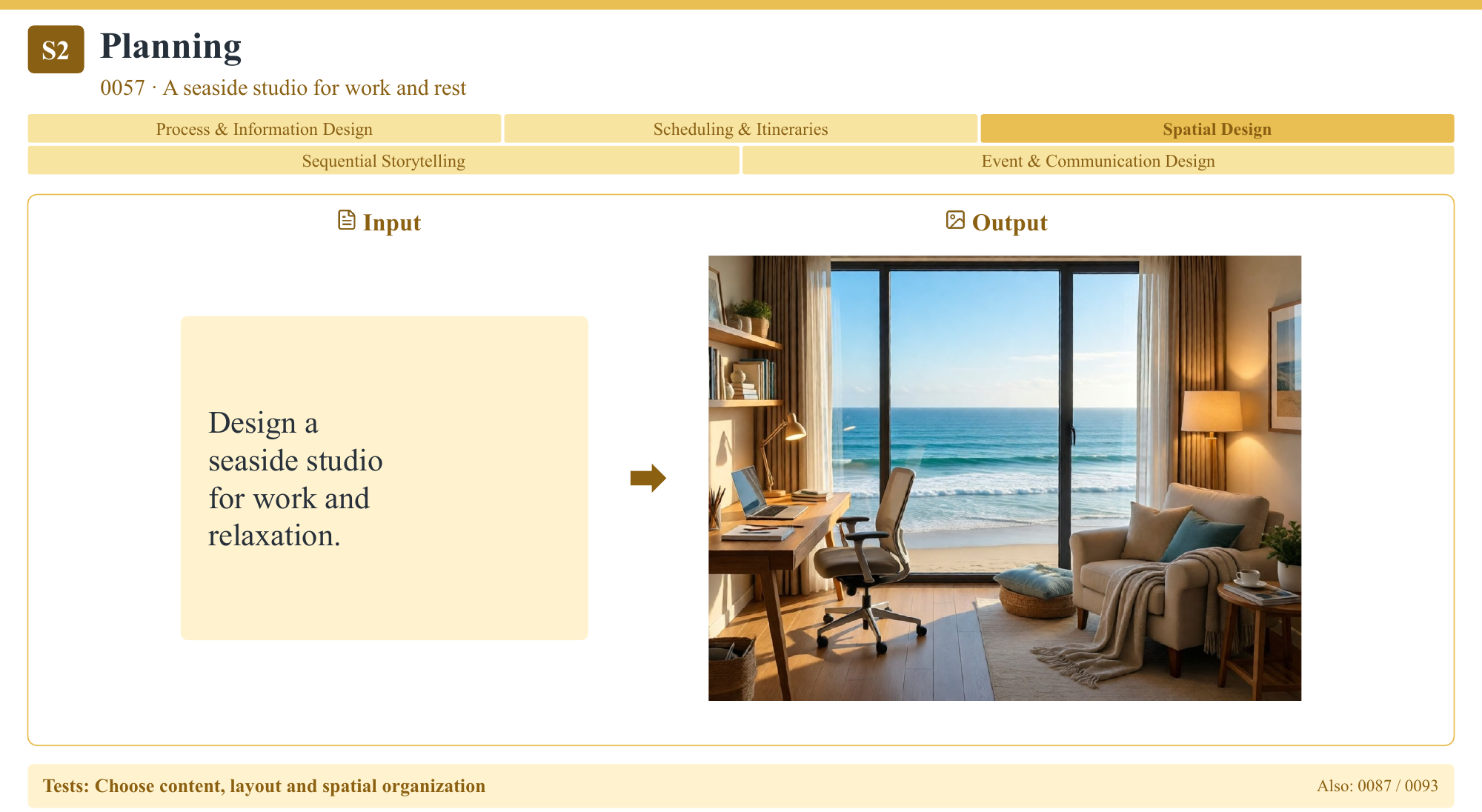}
\caption{\textbf{S2: Planning.} The example translates a request for a seaside studio into a spatial arrangement supporting both work and relaxation.}
\label{fig:sp_task_s2}
\end{figure*}

\begin{figure*}[t]
\centering
\includegraphics[width=\linewidth]
{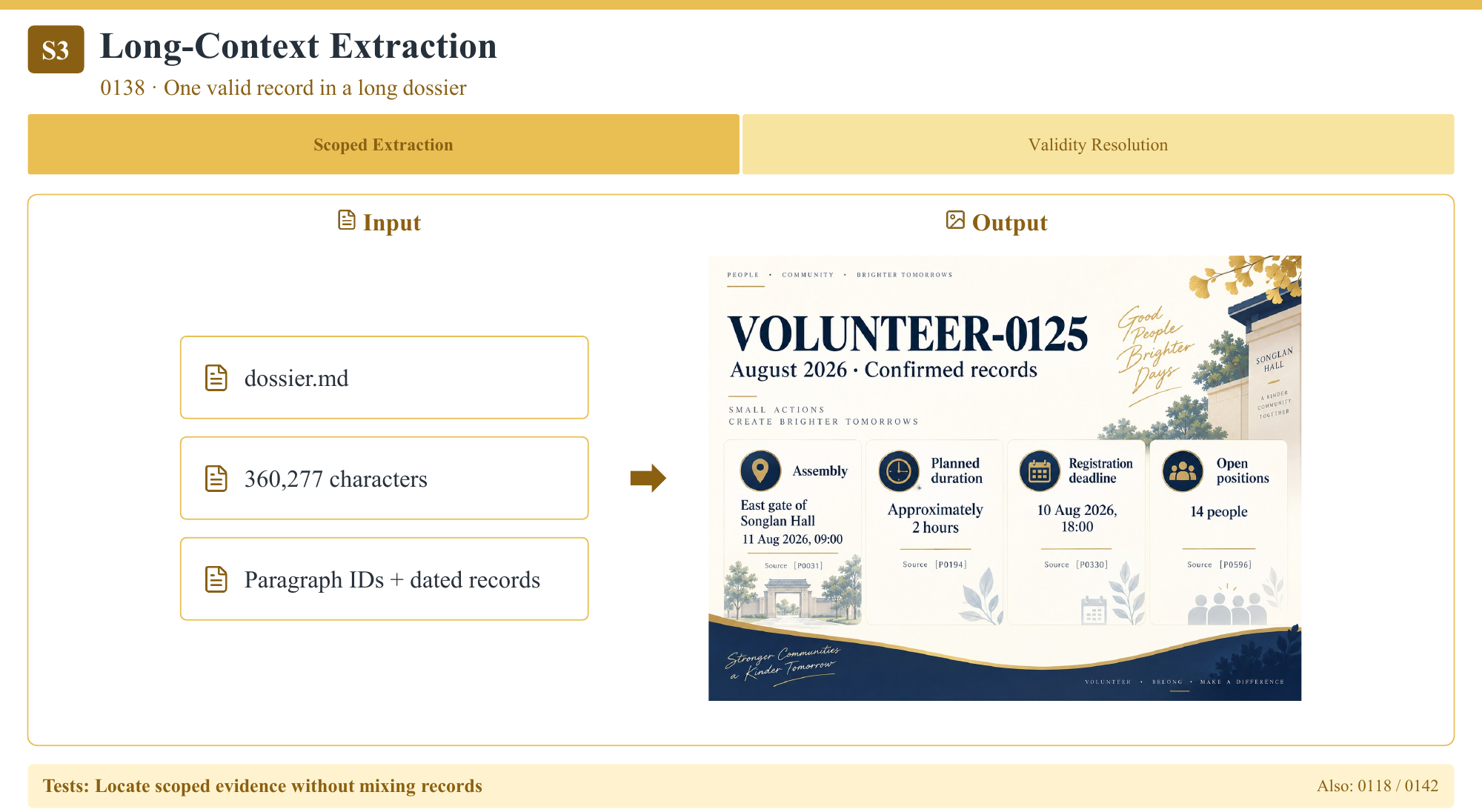}
\caption{\textbf{S3: Long-Context Extraction.} The example extracts the requested record from a long dossier and presents the relevant fields in a visual summary.}
\label{fig:sp_task_s3}
\end{figure*}

\begin{figure*}[t]
\centering
\includegraphics[width=\linewidth]
{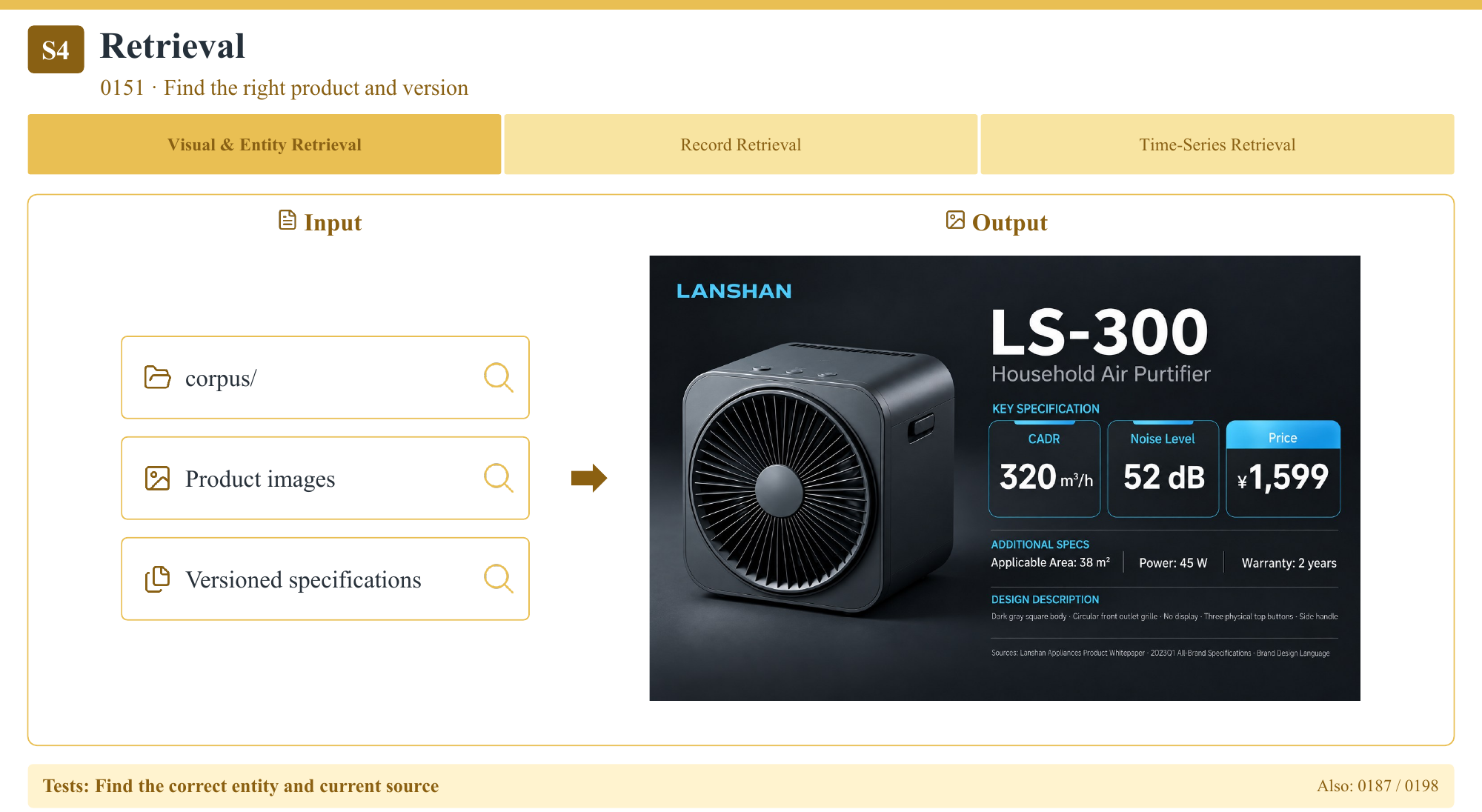}
\caption{\textbf{S4: Retrieval.} The example requires locating the correct product image and current specifications before constructing the visual output.}
\label{fig:sp_task_s4}
\end{figure*}

\begin{figure*}[t]
\centering
\includegraphics[width=\linewidth]
{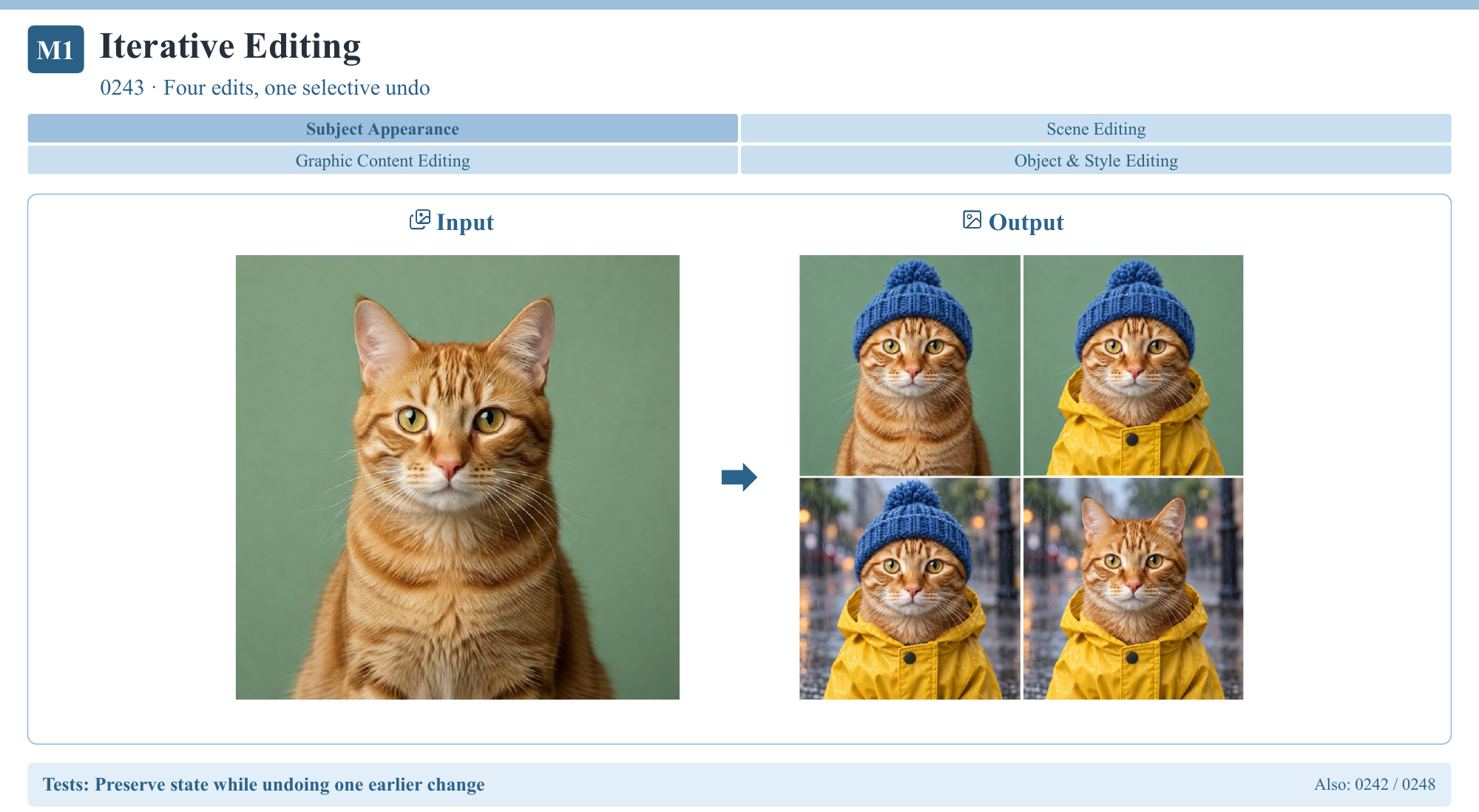}
\caption{\textbf{M1: Iterative Editing.} Successive edits modify the subject and scene. A later instruction reverses one earlier change while retaining the other active edits.}
\label{fig:sp_task_m1}
\end{figure*}

\begin{figure*}[t]
\centering
\includegraphics[width=\linewidth]
{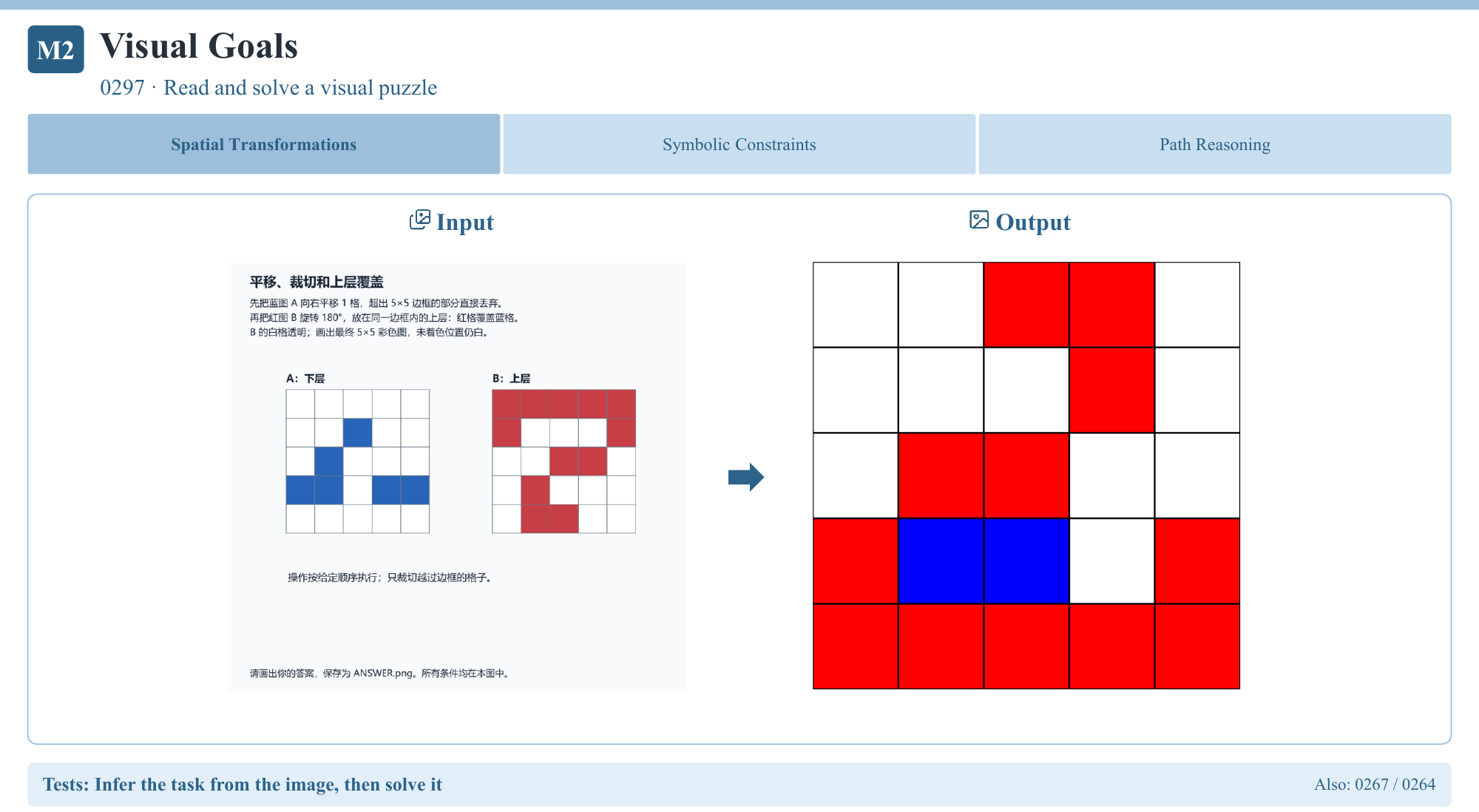}
\caption{\textbf{M2: Visual Goals.} The task instructions and spatial constraints are embedded in the input image. The output visualizes the inferred solution.}
\label{fig:sp_task_m2}
\end{figure*}

\begin{figure*}[t]
\centering
\includegraphics[width=\linewidth]
{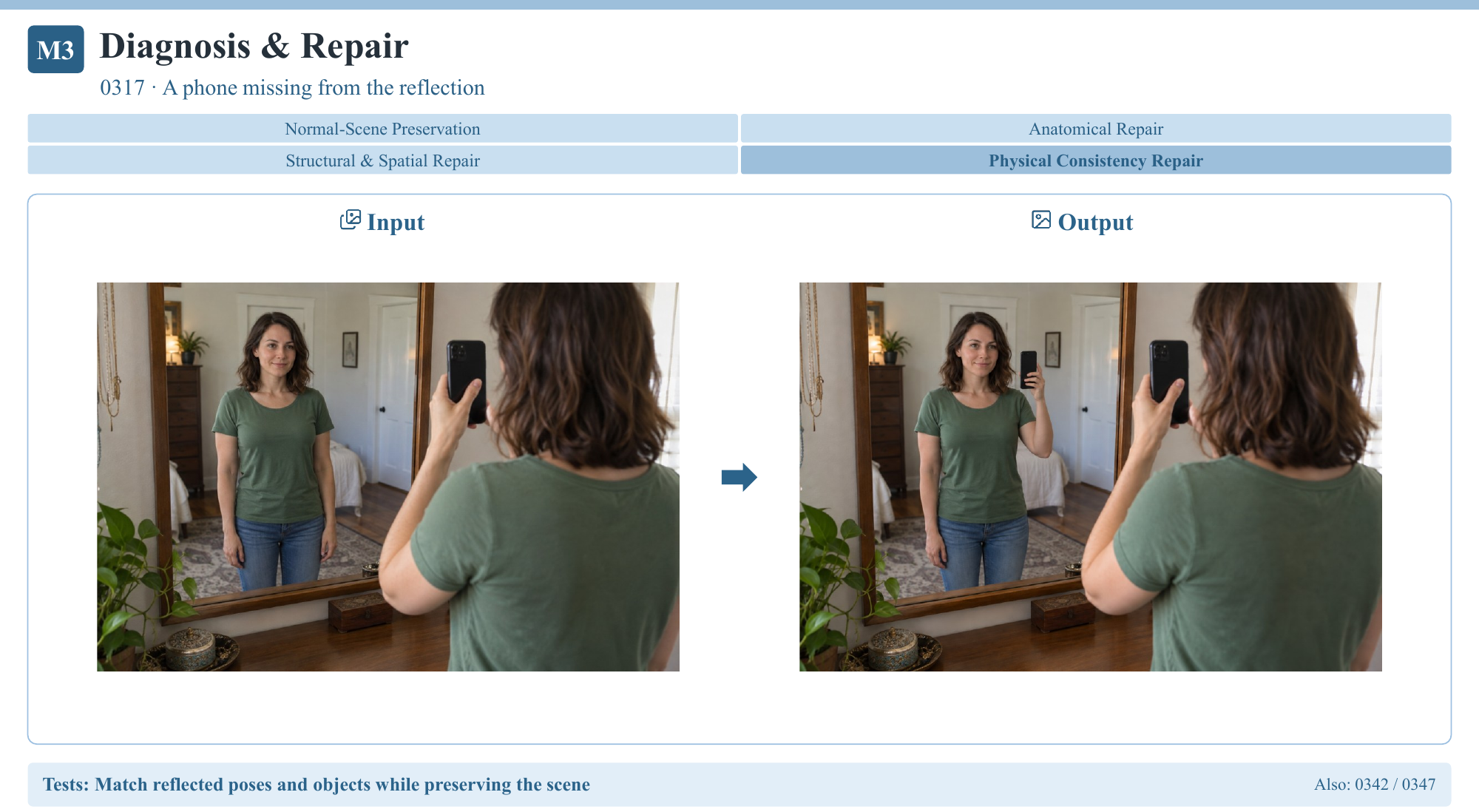}
\caption{\textbf{M3: Diagnosis and Repair.} The example concerns an inconsistent mirror reflection. Repair must address the reflected object while preserving the surrounding scene.}
\label{fig:sp_task_m3}
\end{figure*}

\begin{figure*}[t]
\centering
\includegraphics[width=\linewidth]
{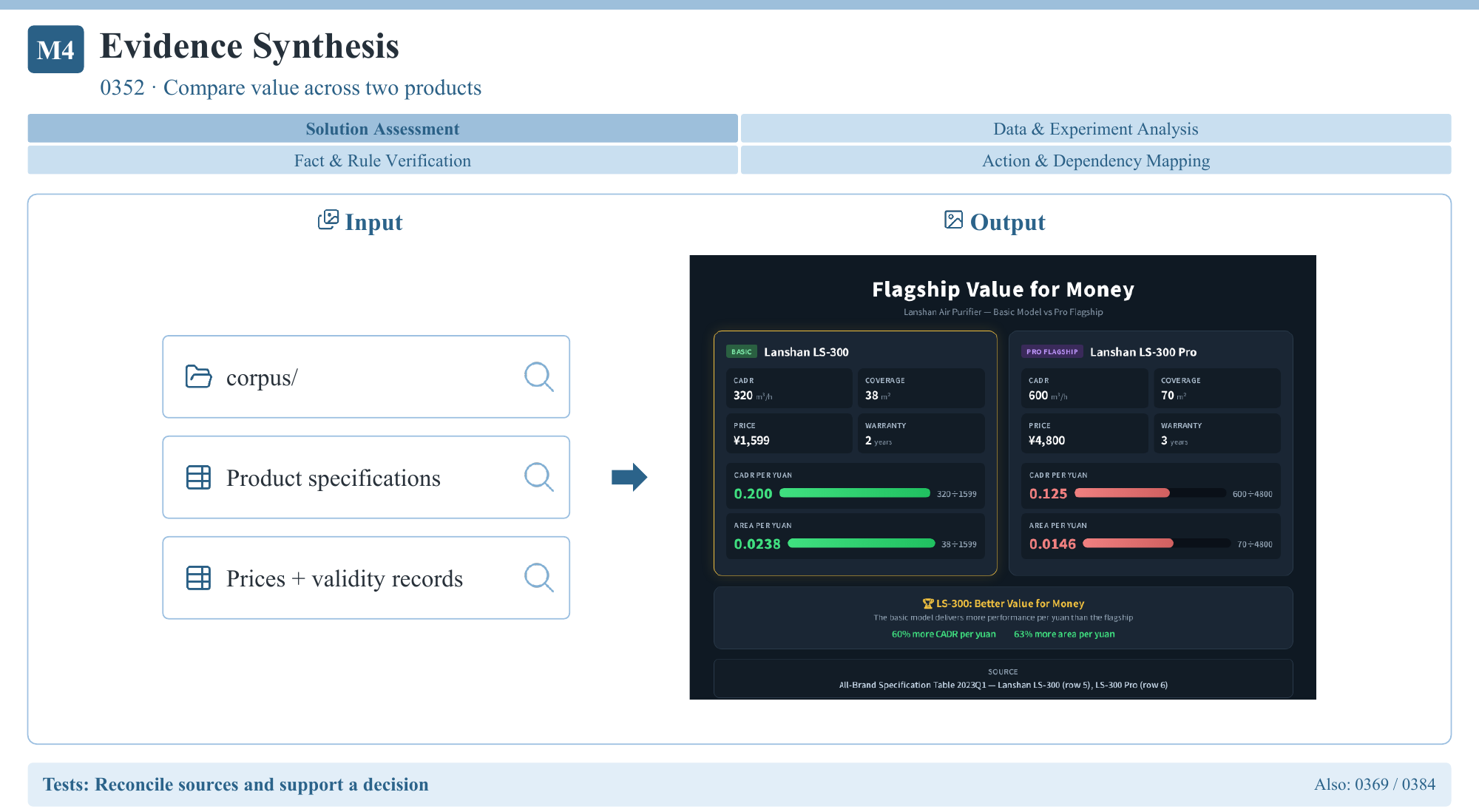}
\caption{\textbf{M4: Evidence Synthesis.} The example combines product specifications, prices, and validity records to construct an evidence-grounded comparison.}
\label{fig:sp_task_m4}
\end{figure*}

\begin{figure*}[t]
\centering
\includegraphics[width=\linewidth]
{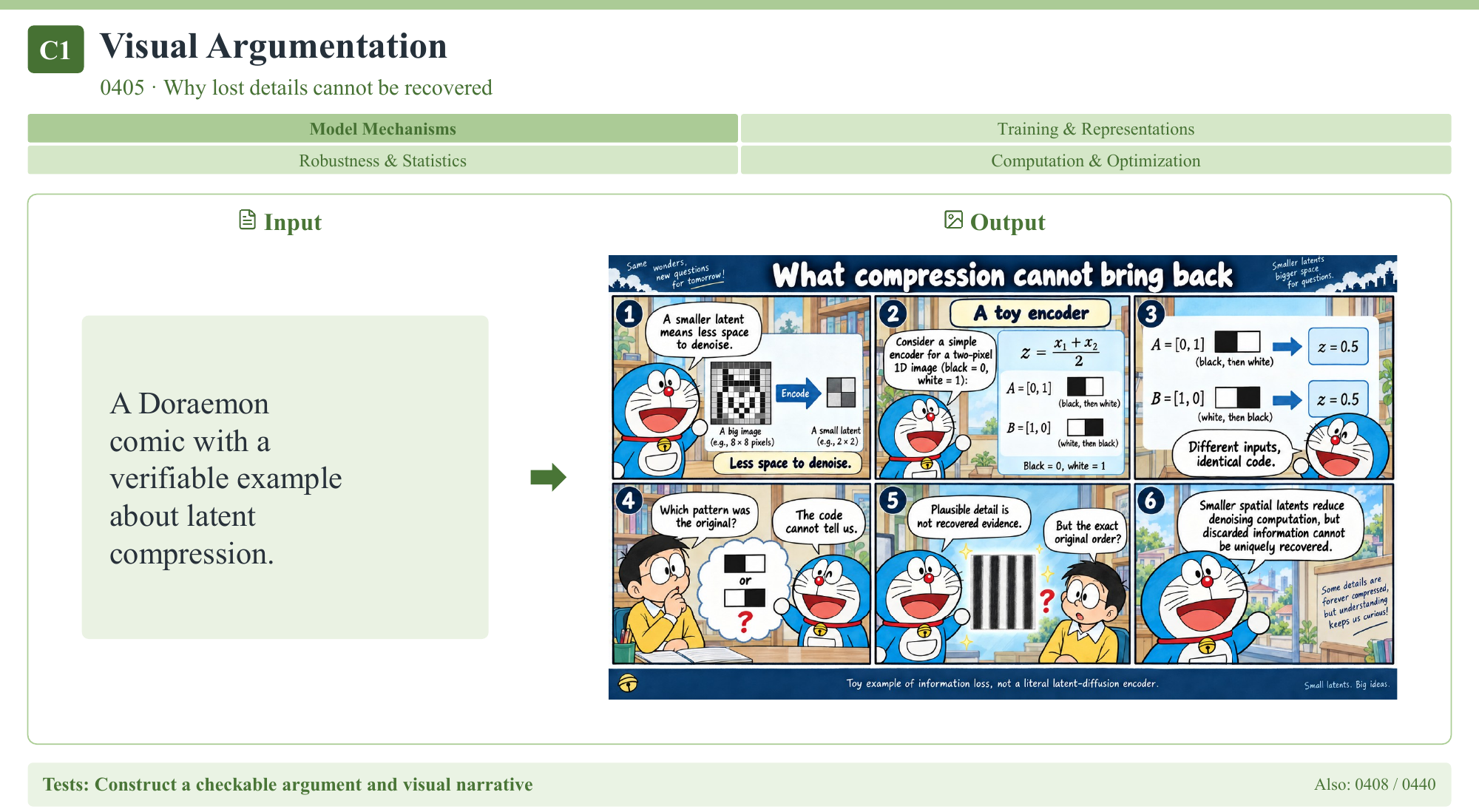}
\caption{\textbf{C1: Visual Argumentation.} The example uses a constructed case and a comic narrative to explain why compression can prevent recovery of the original information.}
\label{fig:sp_task_c1}
\end{figure*}

\begin{figure*}[t]
\centering
\includegraphics[width=\linewidth]
{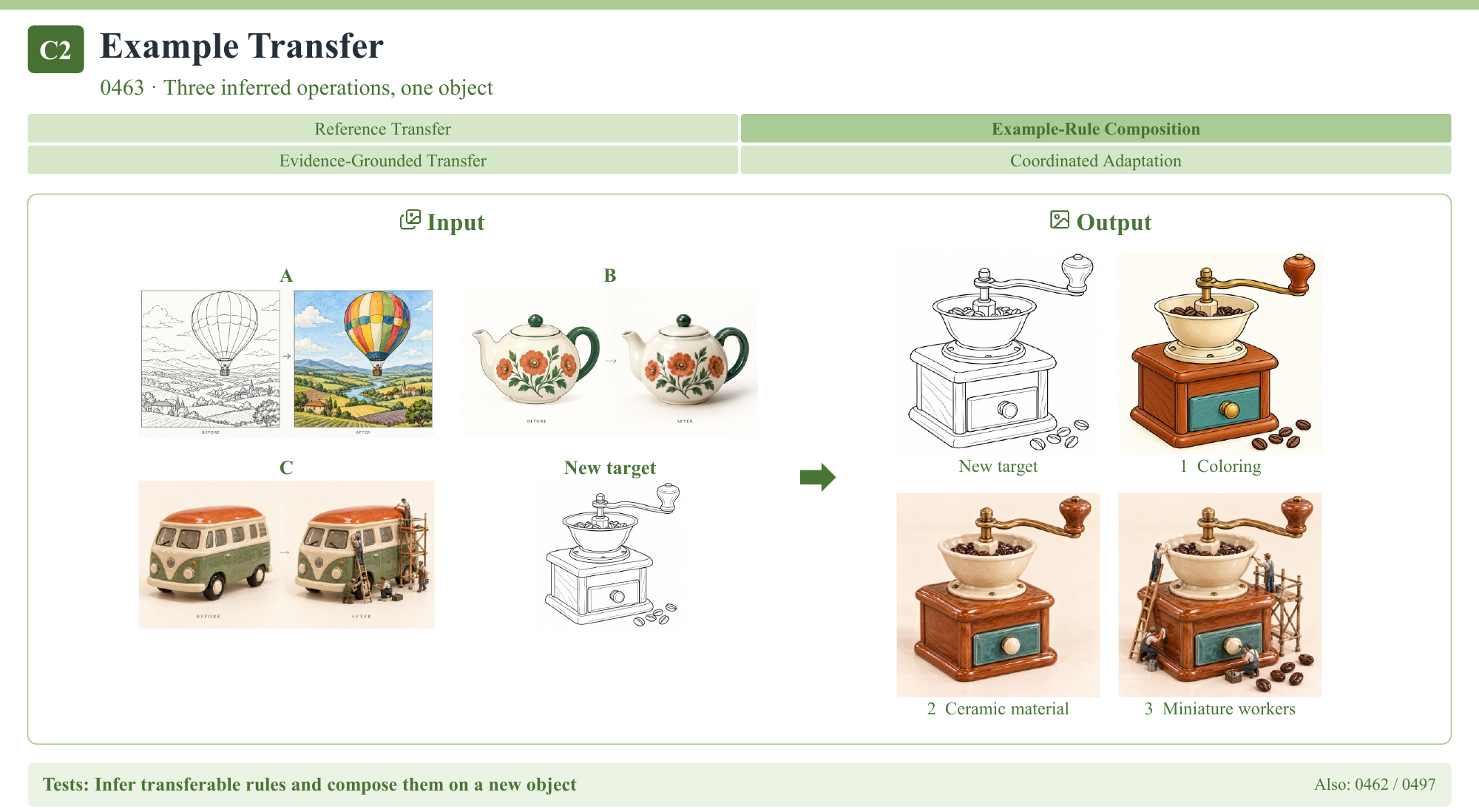}
\caption{\textbf{C2: Example Transfer.} Three paired demonstrations define coloring, ceramic material conversion, and miniature-worker composition. The illustrated sequence applies these operations to a new object.}
\label{fig:sp_task_c2}
\end{figure*}

\begin{figure*}[t]
\centering
\includegraphics[width=\linewidth]
{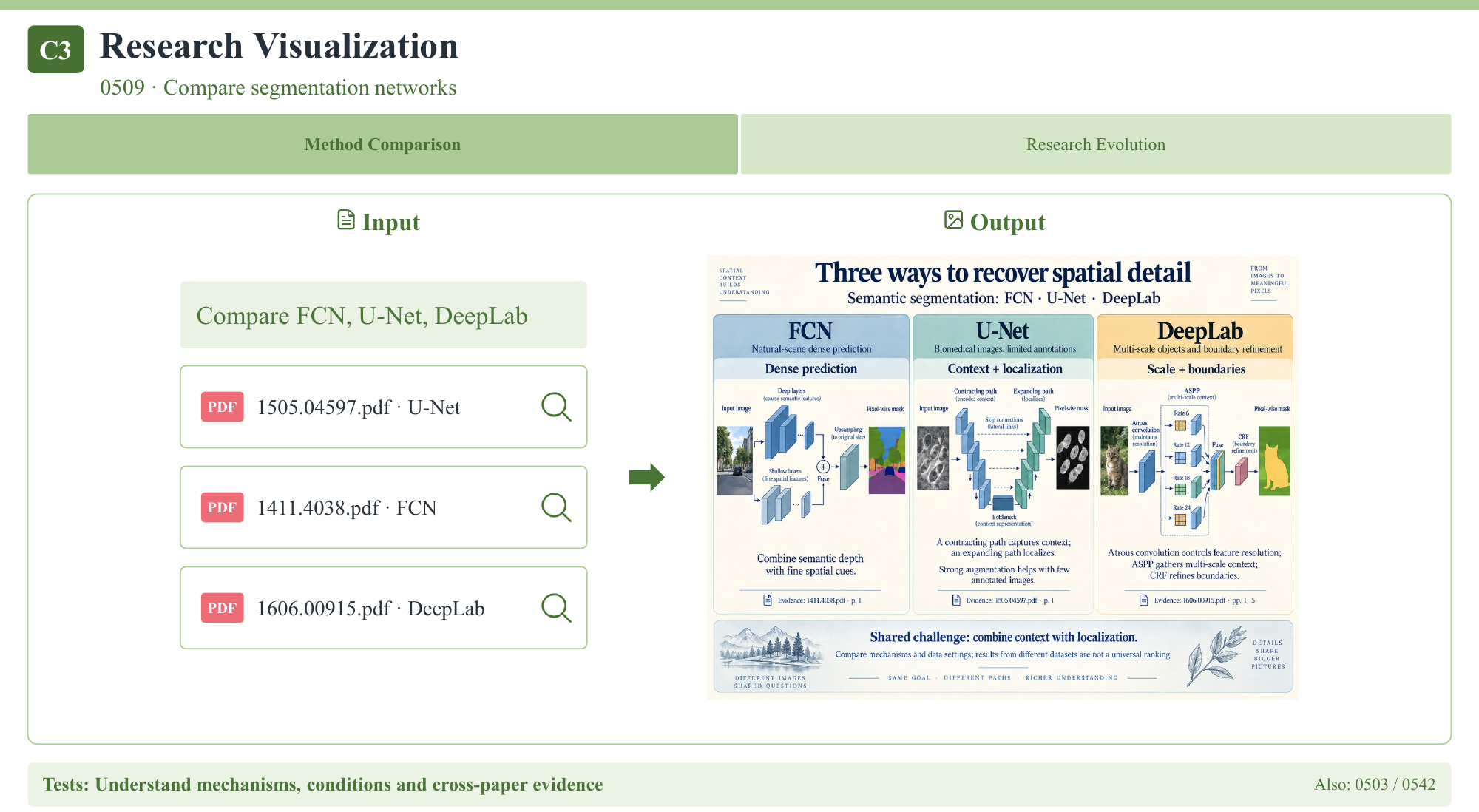}
\caption{\textbf{C3: Scientific Visualization.} The example integrates supplied papers into a visual comparison of FCN, U-Net, and DeepLab, including their approaches to recovering spatial detail.}
\label{fig:sp_task_c3}
\end{figure*}

\begin{figure*}[t]
\centering
\includegraphics[width=\linewidth]
{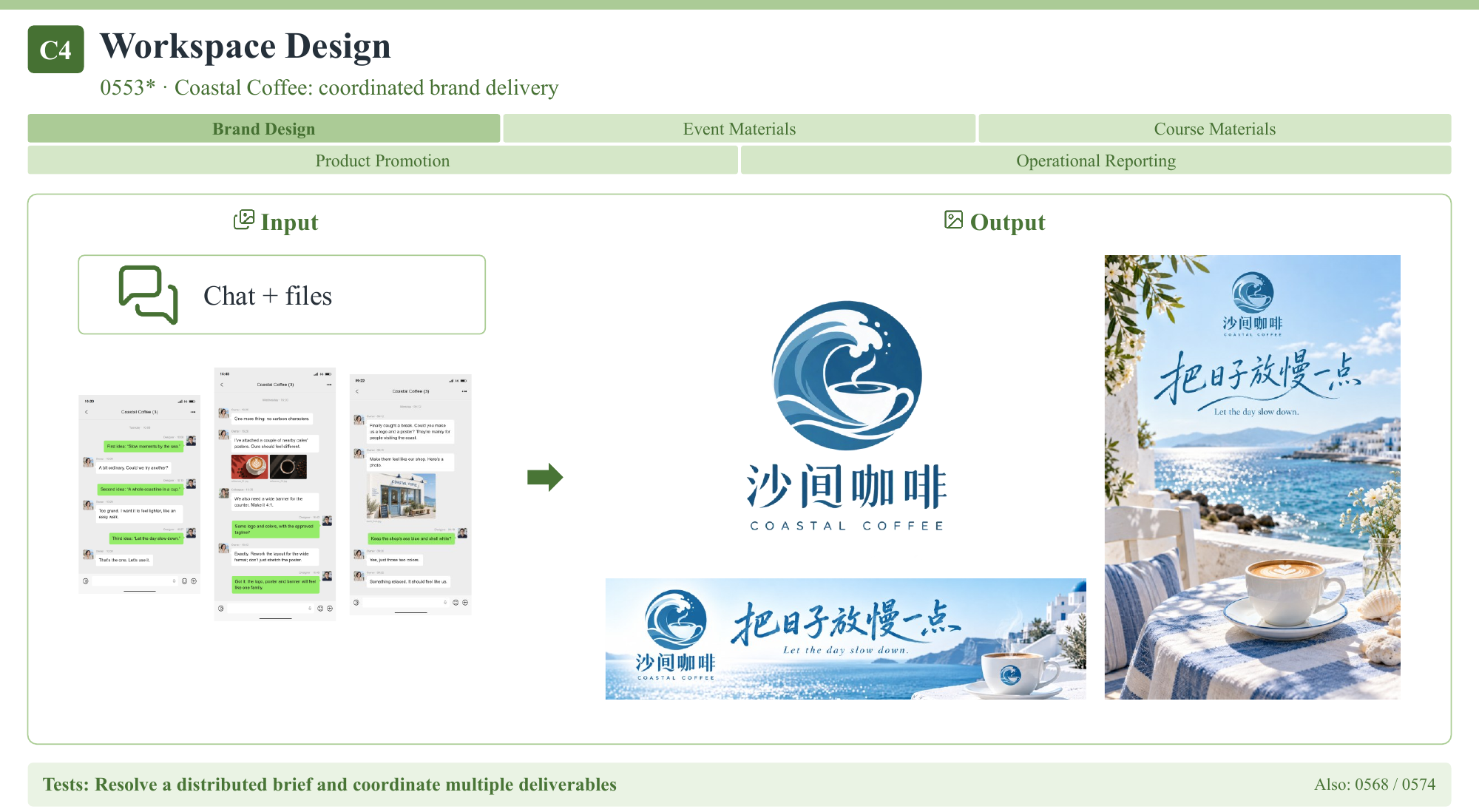}
\caption{\textbf{C4: Workspace Design.} The example combines conversation history and source files into a coordinated coffee-shop identity, poster, and banner.}
\label{fig:sp_task_c4}
\end{figure*}
\clearpage
\subsection{Comparison with Existing Systems}
\label{app:visual_comparison}

\begin{figure*}[t]
\centering
\includegraphics[page=1,width=\linewidth]
{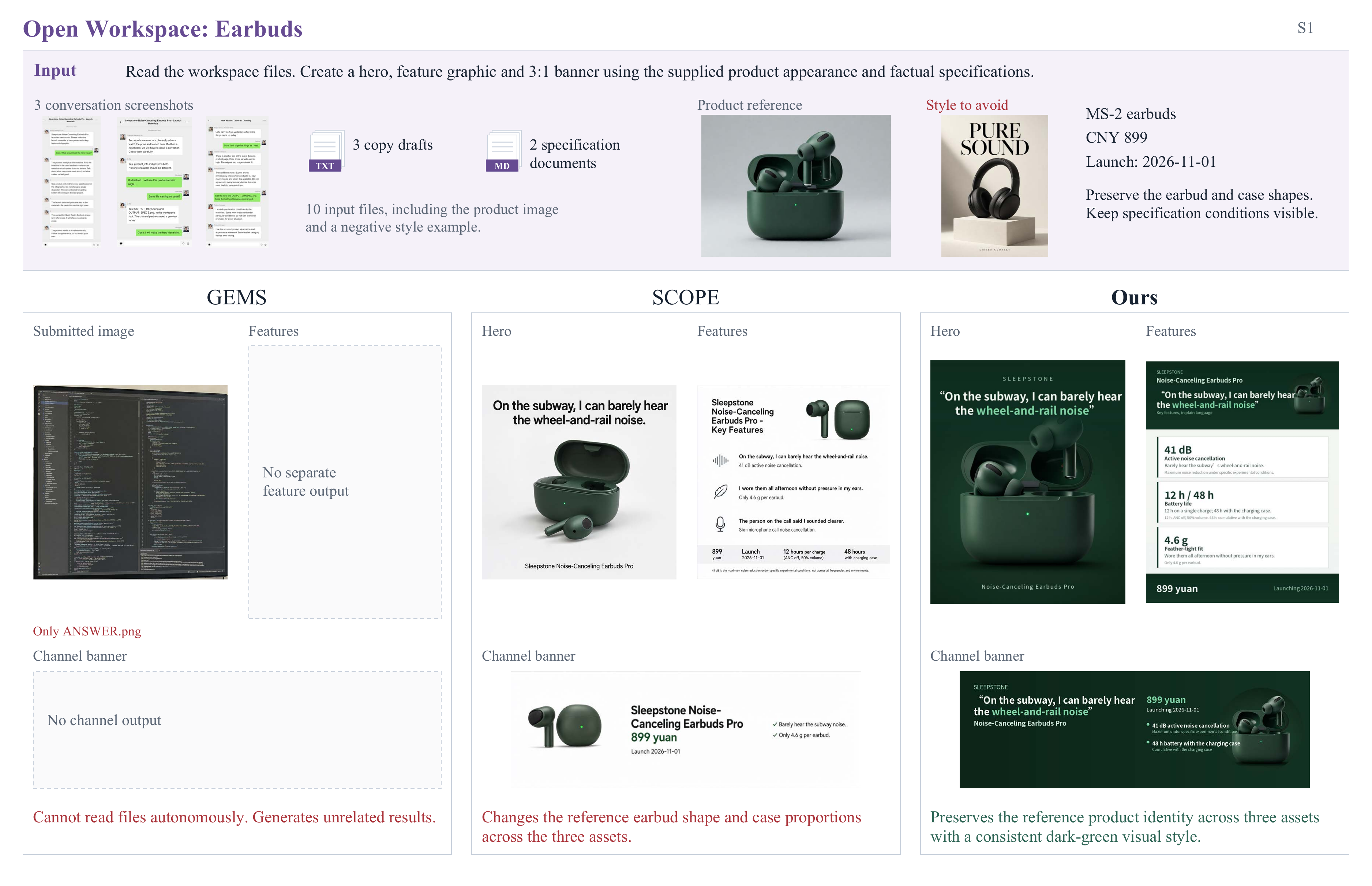}
\caption{\textbf{Open workspace: earbuds.}
Inputs are shown above; outputs from GEMS, SCOPE, and
V-Protocol (Ours) are shown below.
Empty slots indicate missing deliverables.}
\label{fig:sp_comparison}
\end{figure*}

\begin{figure*}[t]
\centering
\includegraphics[page=2,width=\linewidth]
{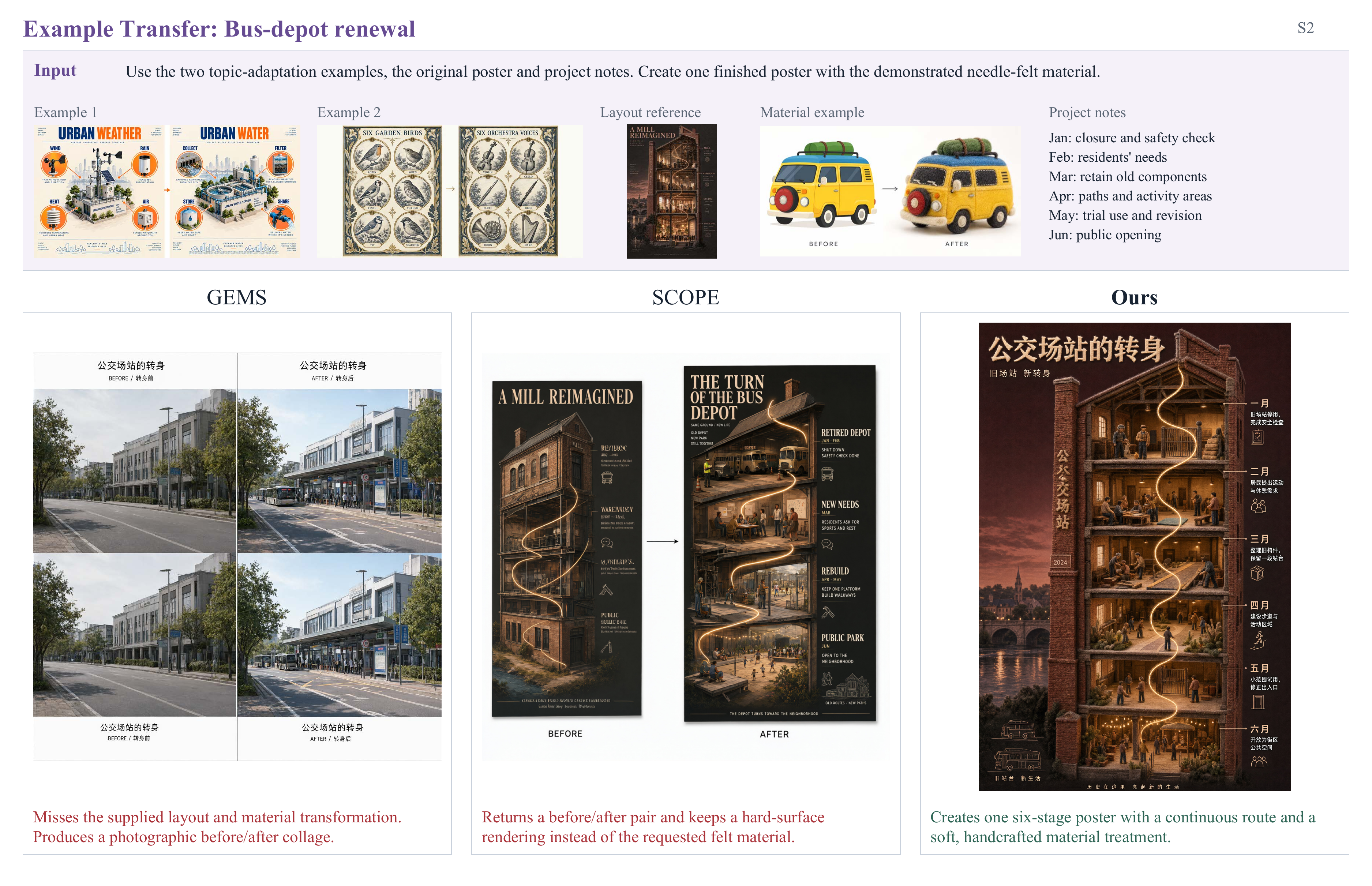}
\caption{\textbf{Example transfer: bus-depot renewal.}
Comparison of topic adaptation and material transformation.}
\label{fig:sp_comparison_transfer}
\end{figure*}

\begin{figure*}[t]
\centering
\includegraphics[page=3,width=\linewidth]
{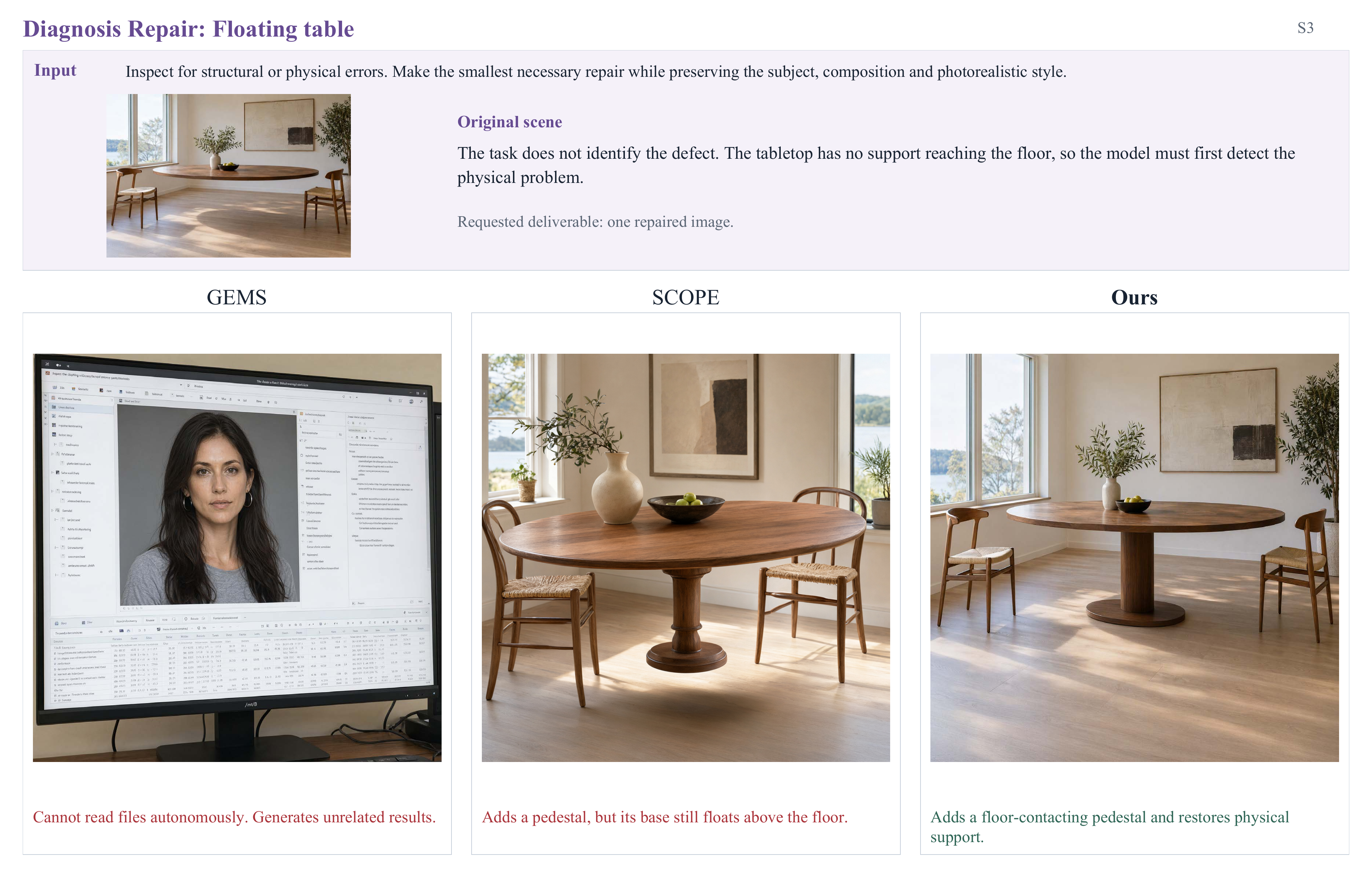}
\caption{\textbf{Diagnosis and repair: floating table.}
Comparison of physical-error detection and repair.}
\label{fig:sp_comparison_repair}
\end{figure*}

\begin{figure*}[t]
\centering
\includegraphics[page=4,width=\linewidth]
{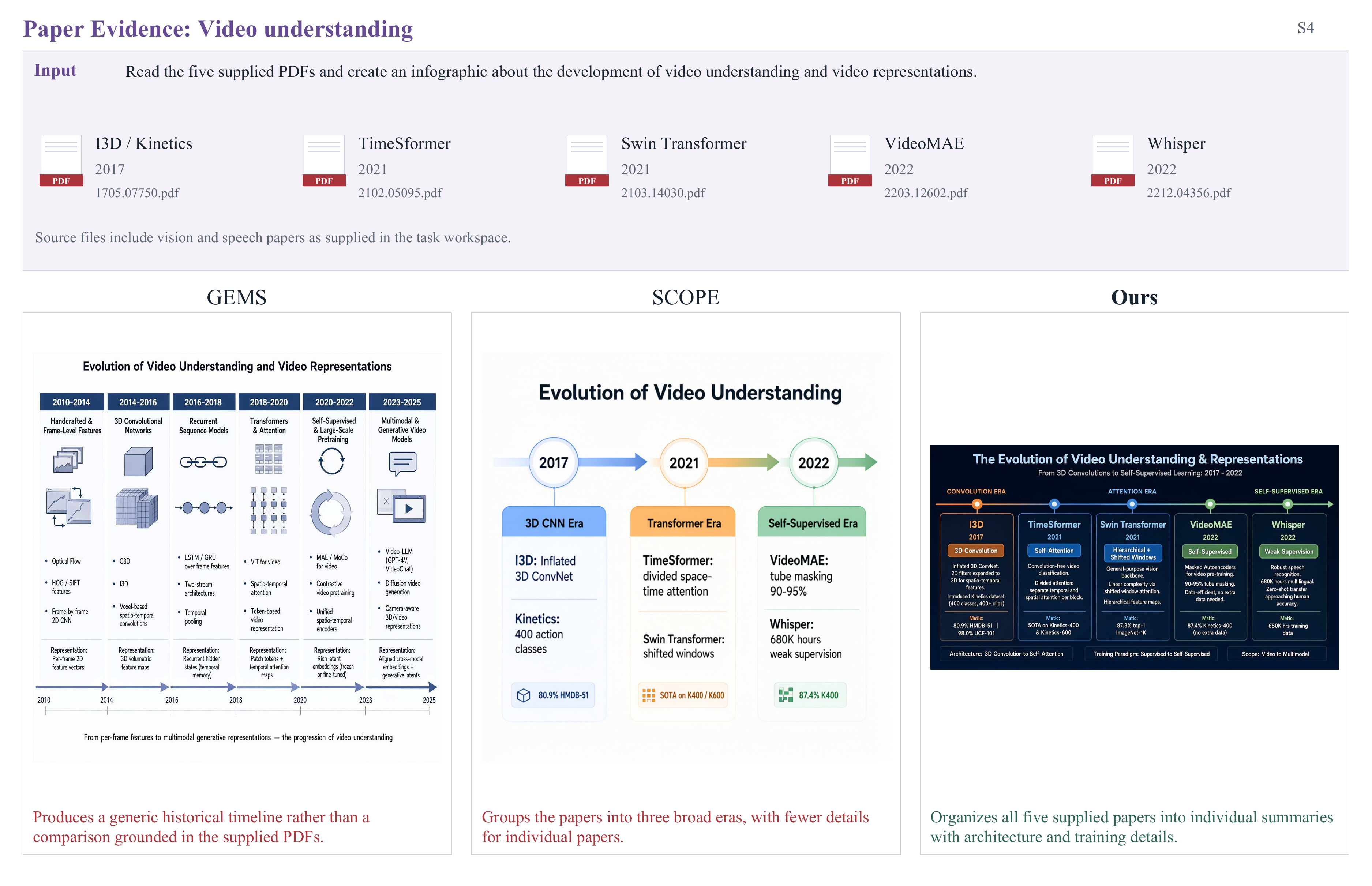}
\caption{\textbf{Paper evidence: video understanding.}
Comparison of evidence coverage and information organization
across the supplied papers.}
\label{fig:sp_comparison_video}
\end{figure*}

Figures~\ref{fig:sp_comparison}--\ref{fig:sp_comparison_video}
present selected outputs from GEMS, SCOPE, and V-Protocol
on the same tasks. These qualitative cases illustrate
differences in transformation realization, evidence coverage,
and coordinated delivery.

\clearpage
\subsection{Additional Visual Gallery}
\label{app:visual_gallery}

The gallery presents selected visual examples grouped by theme,
including identity, layout, illustration, and scene composition.
These examples provide visual context and are mainly from the
quantitative benchmark comparisons.

\begin{figure*}[t]
\centering
\includegraphics[width=\linewidth]
{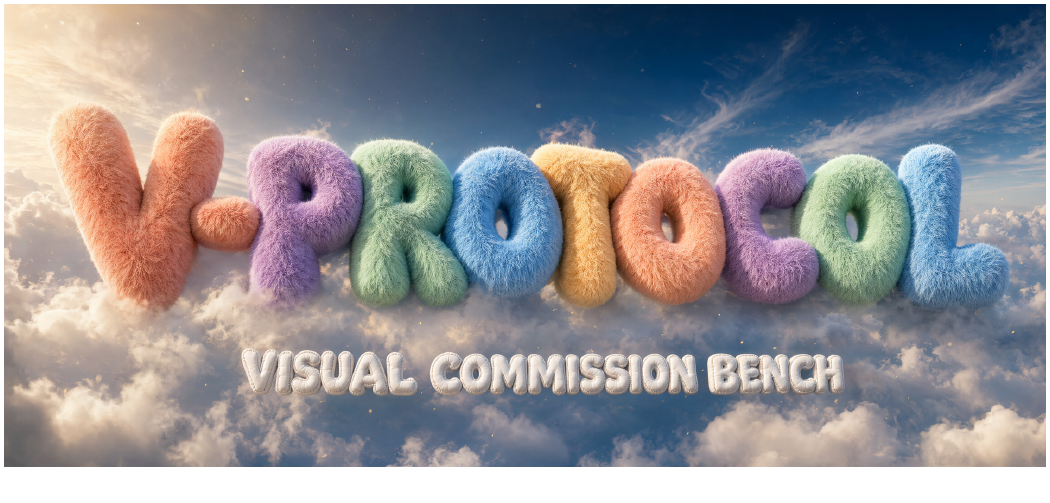}
\caption{\textbf{Qualitative gallery: visual identity.} Typography, material, and composition in the V-Protocol identity illustration.}
\label{fig:sp_g1}
\end{figure*}

\begin{figure*}[t]
\centering
\includegraphics[width=\linewidth]
{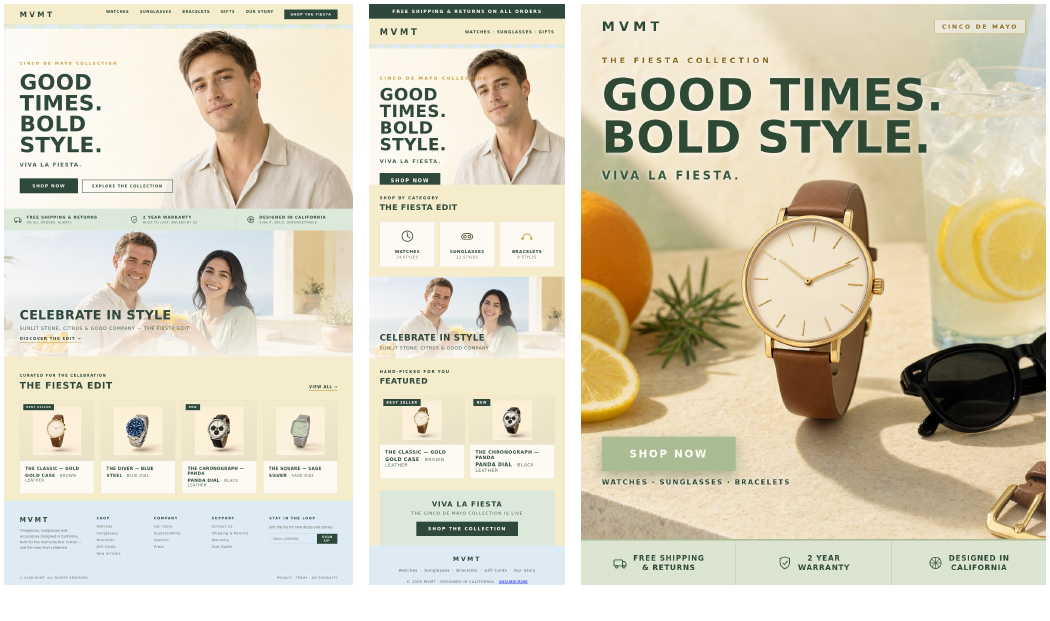}
\caption{\textbf{Qualitative gallery: brand and layout adaptation.} Coordinated brand presentation across different page formats.}
\label{fig:sp_g2}
\end{figure*}

\begin{figure*}[t]
\centering
\includegraphics[width=\linewidth]
{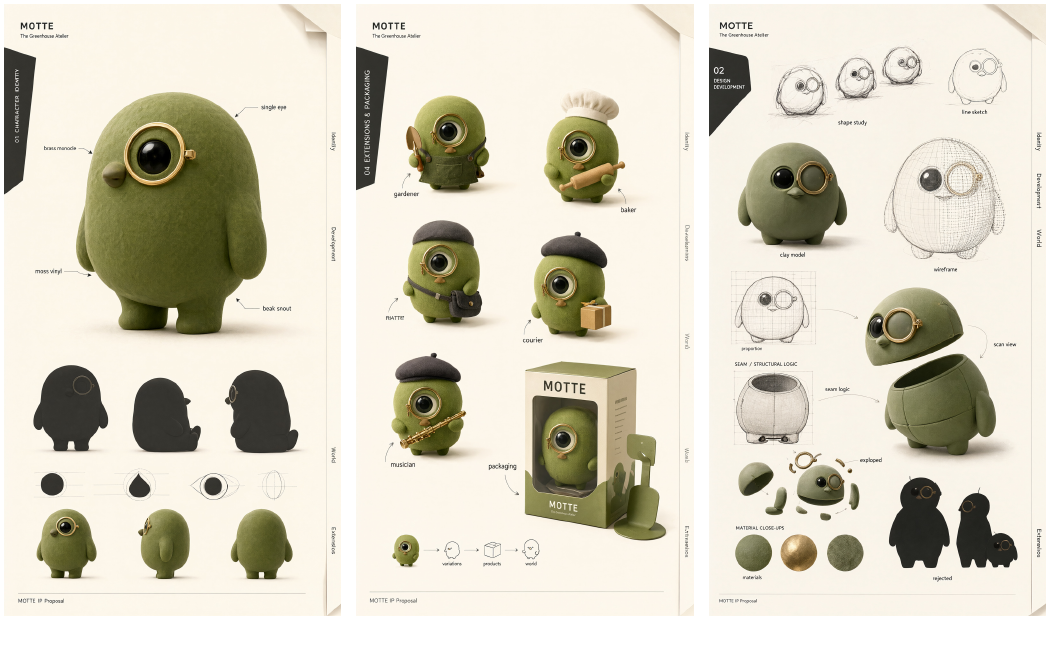}
\caption{\textbf{Qualitative gallery: character design.} Character appearance, multiple views, and related presentation materials.}
\label{fig:sp_g3}
\end{figure*}

\begin{figure*}[t]
\centering
\includegraphics[width=\linewidth]
{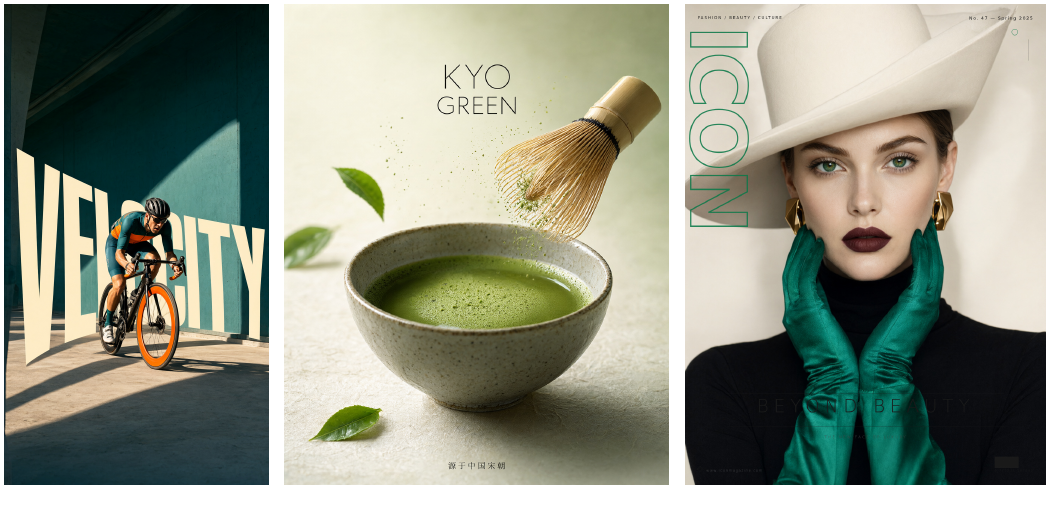}
\caption{\textbf{Qualitative gallery: poster design.} Examples combining imagery, typography, and promotional composition.}
\label{fig:sp_g4}
\end{figure*}

\begin{figure*}[t]
\centering
\includegraphics[width=\linewidth]
{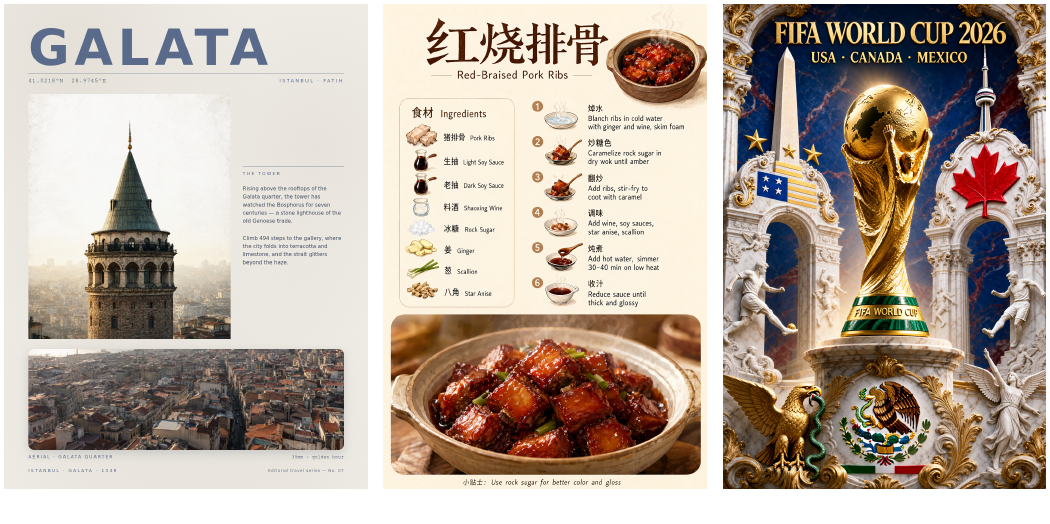}
\caption{\textbf{Qualitative gallery: visual information.} Examples organizing explanatory and promotional content through text and imagery.}
\label{fig:sp_g5}
\end{figure*}

\begin{figure*}[t]
\centering
\includegraphics[width=\linewidth,height=0.78\textheight,keepaspectratio]
{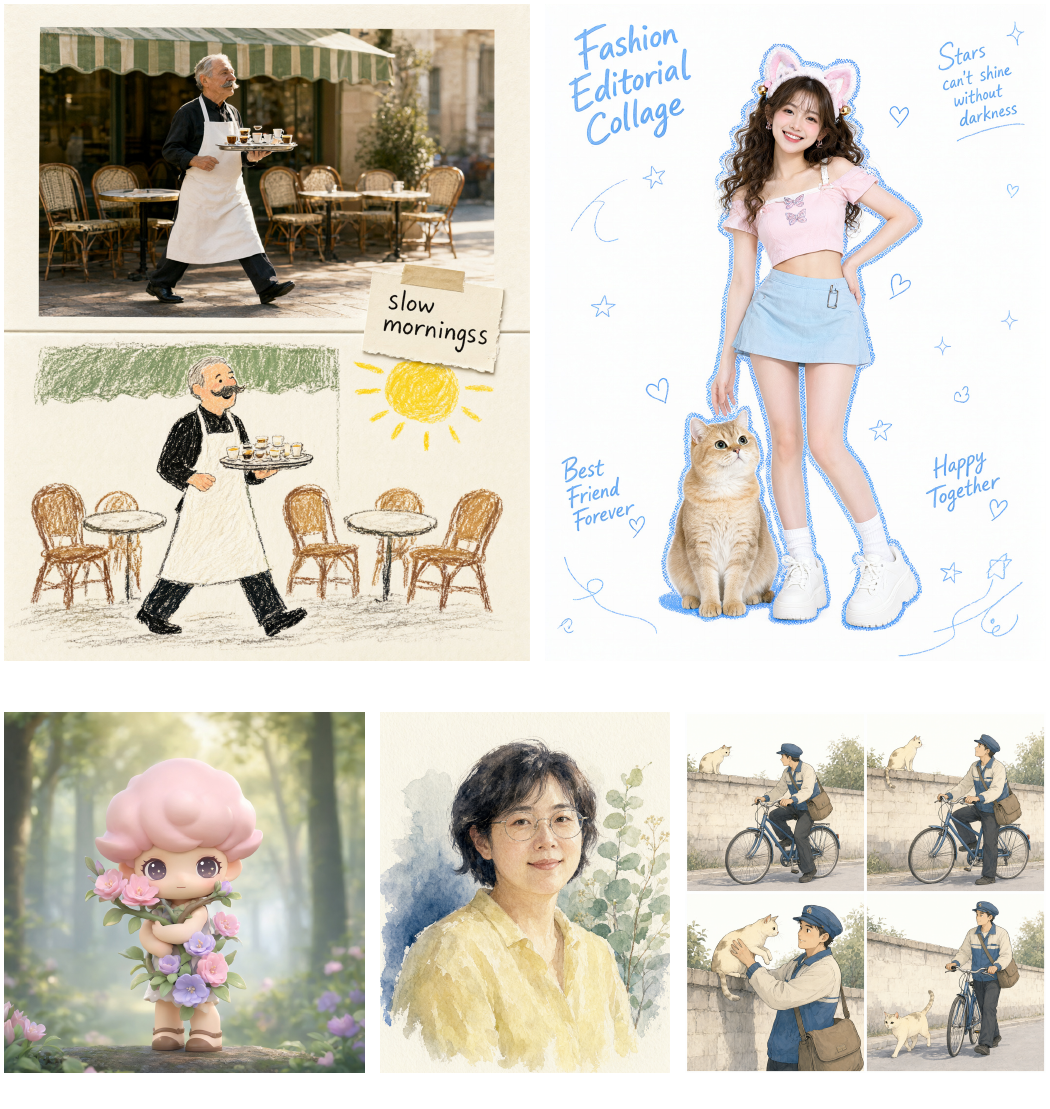}
\caption{\textbf{Qualitative gallery: illustration and visual storytelling.} Examples of stylized depiction and narrative composition.}
\label{fig:sp_g6}
\end{figure*}

\begin{figure*}[t]
\centering
\includegraphics[width=\linewidth]
{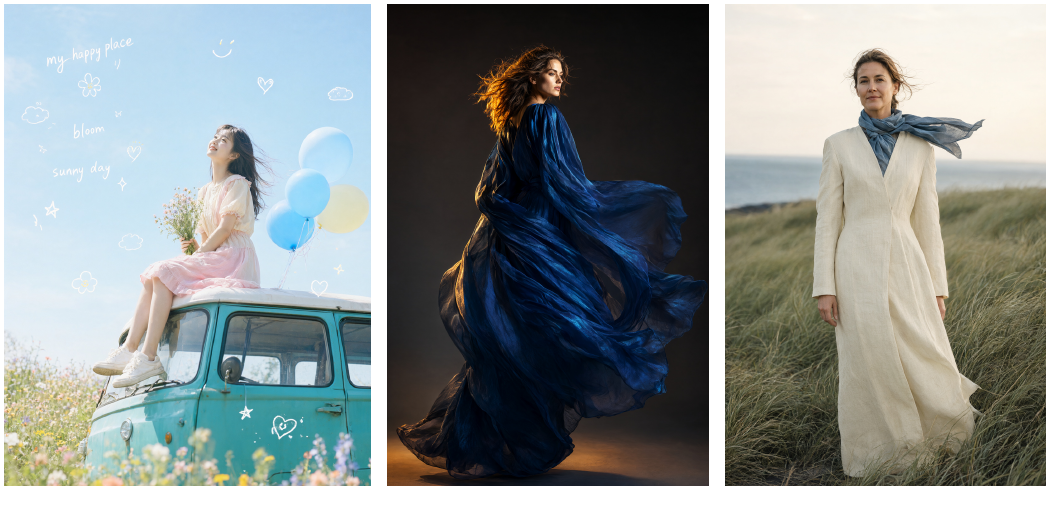}
\caption{\textbf{Qualitative gallery: fashion and portraits.} Examples of portrait treatment, styling, and editorial composition.}
\label{fig:sp_g7}
\end{figure*}

\begin{figure*}[t]
\centering
\includegraphics[width=\linewidth,height=0.78\textheight,keepaspectratio]
{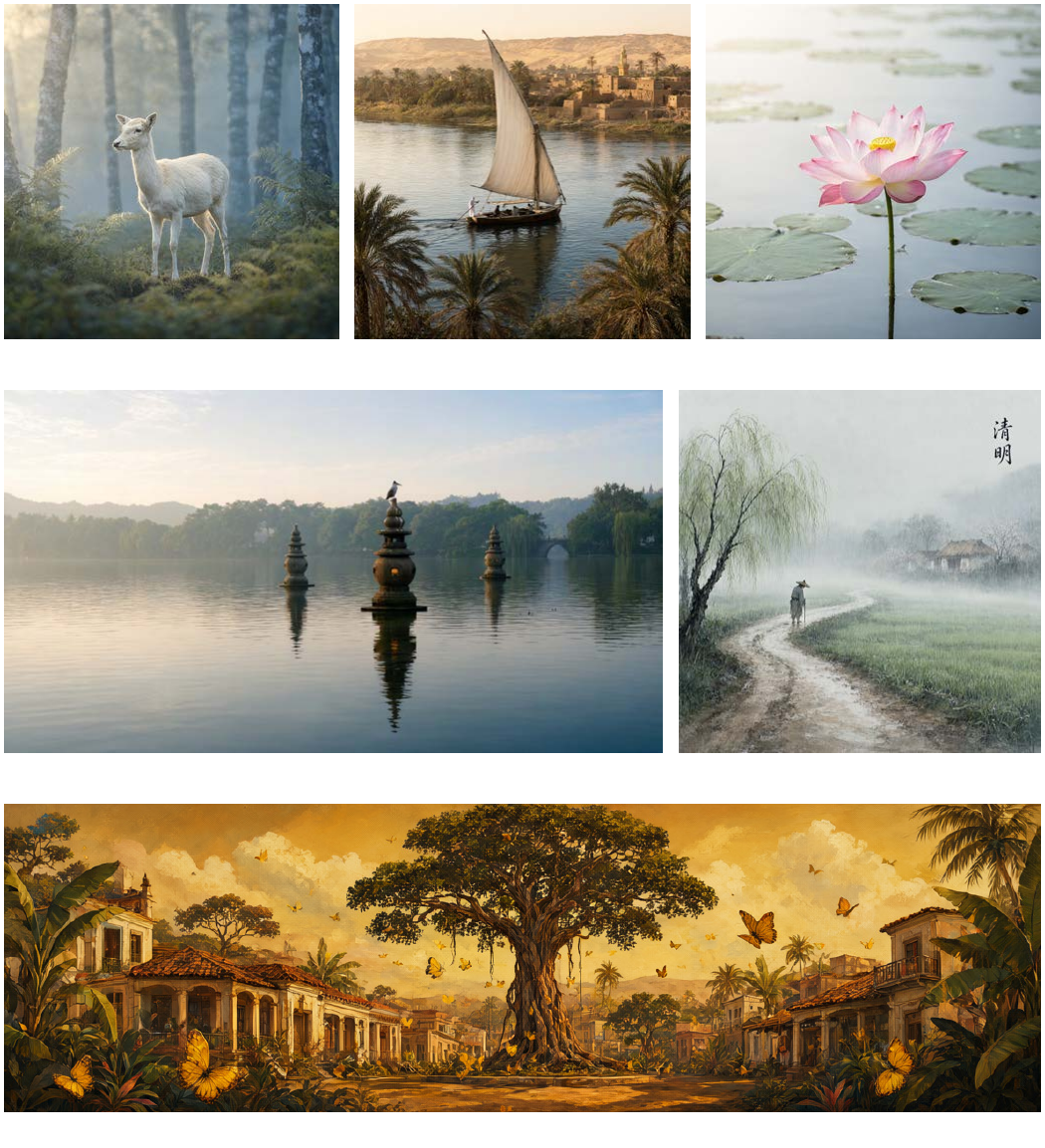}
\caption{\textbf{Qualitative gallery: landscapes and atmosphere.} Examples varying scene composition, lighting, and environmental detail.}
\label{fig:sp_g8}
\end{figure*}

\begin{figure*}[t]
\centering
\includegraphics[width=\linewidth]
{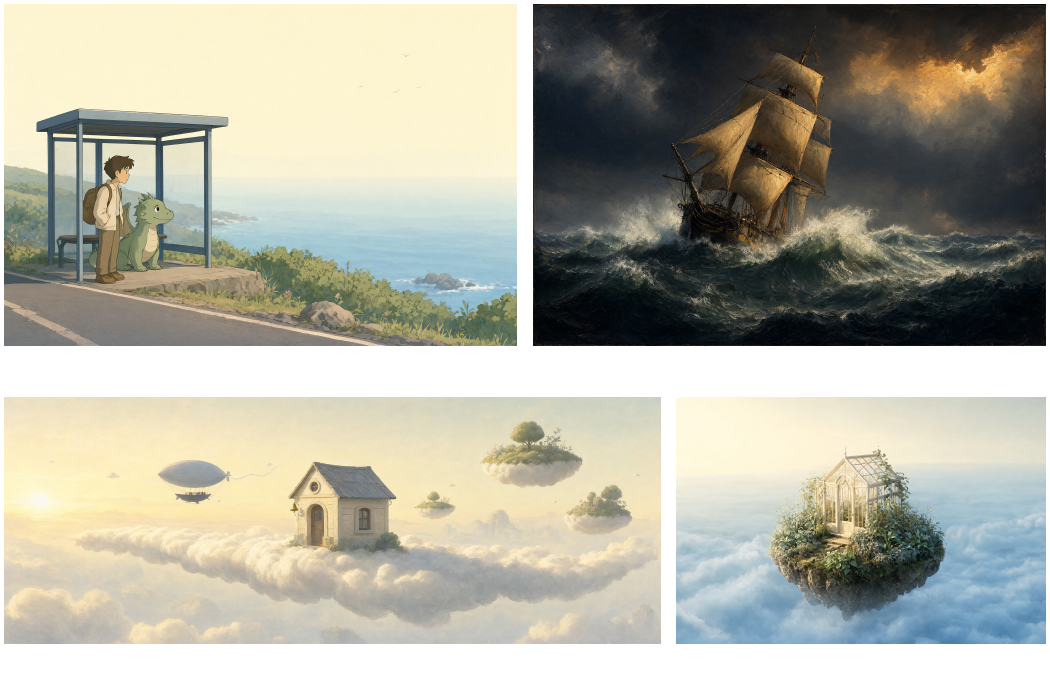}
\caption{\textbf{Qualitative gallery: imagined scenes.} Examples combining familiar subjects with constructed visual settings.}
\label{fig:sp_g9}
\end{figure*}

\begin{figure*}[t]
\centering
\includegraphics[width=\linewidth,height=0.78\textheight,keepaspectratio]
{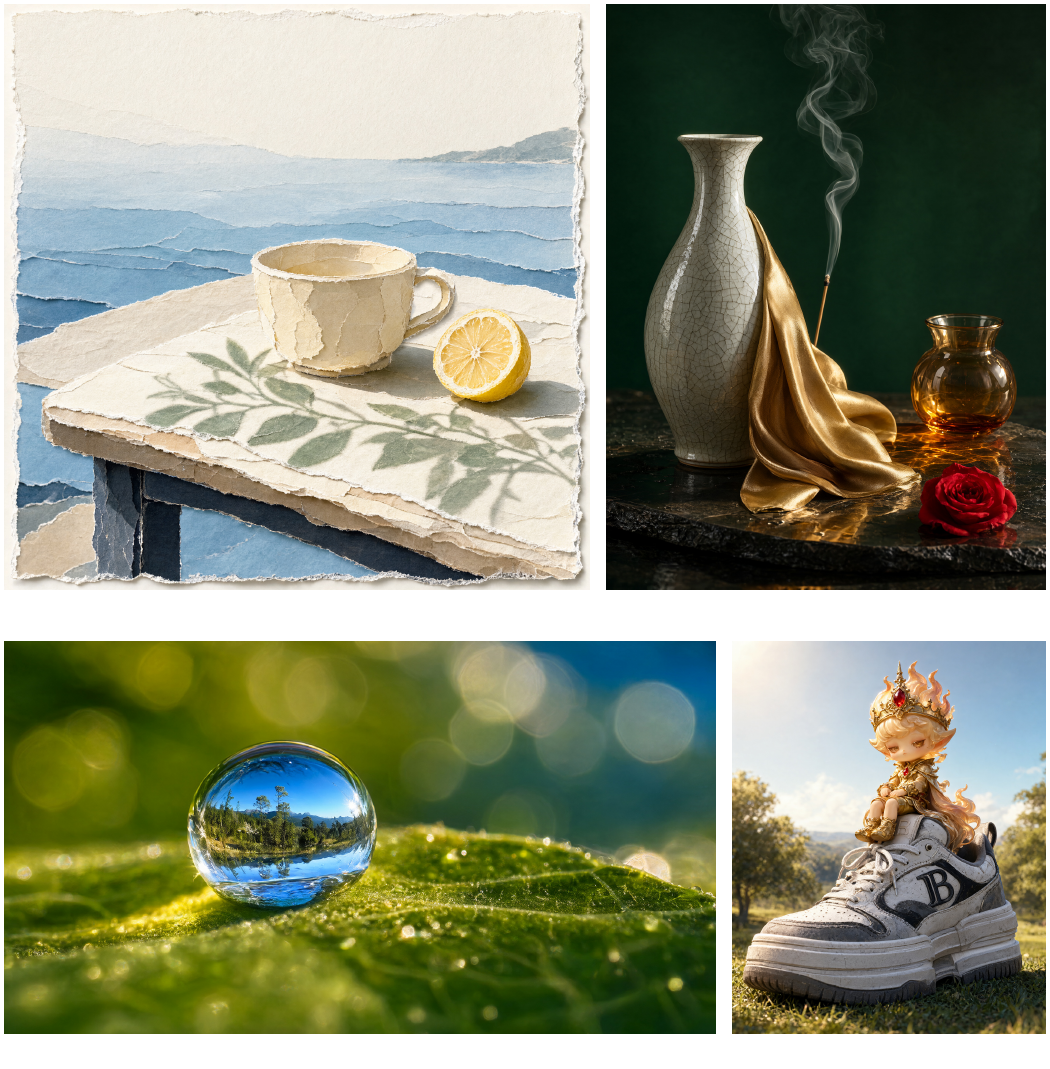}
\caption{\textbf{Qualitative gallery: material and detail.} Examples emphasizing surface appearance, texture, and object composition.}
\label{fig:sp_g10}
\end{figure*}

\begin{figure*}[t]
\centering
\includegraphics[width=\linewidth,height=0.78\textheight,keepaspectratio]
{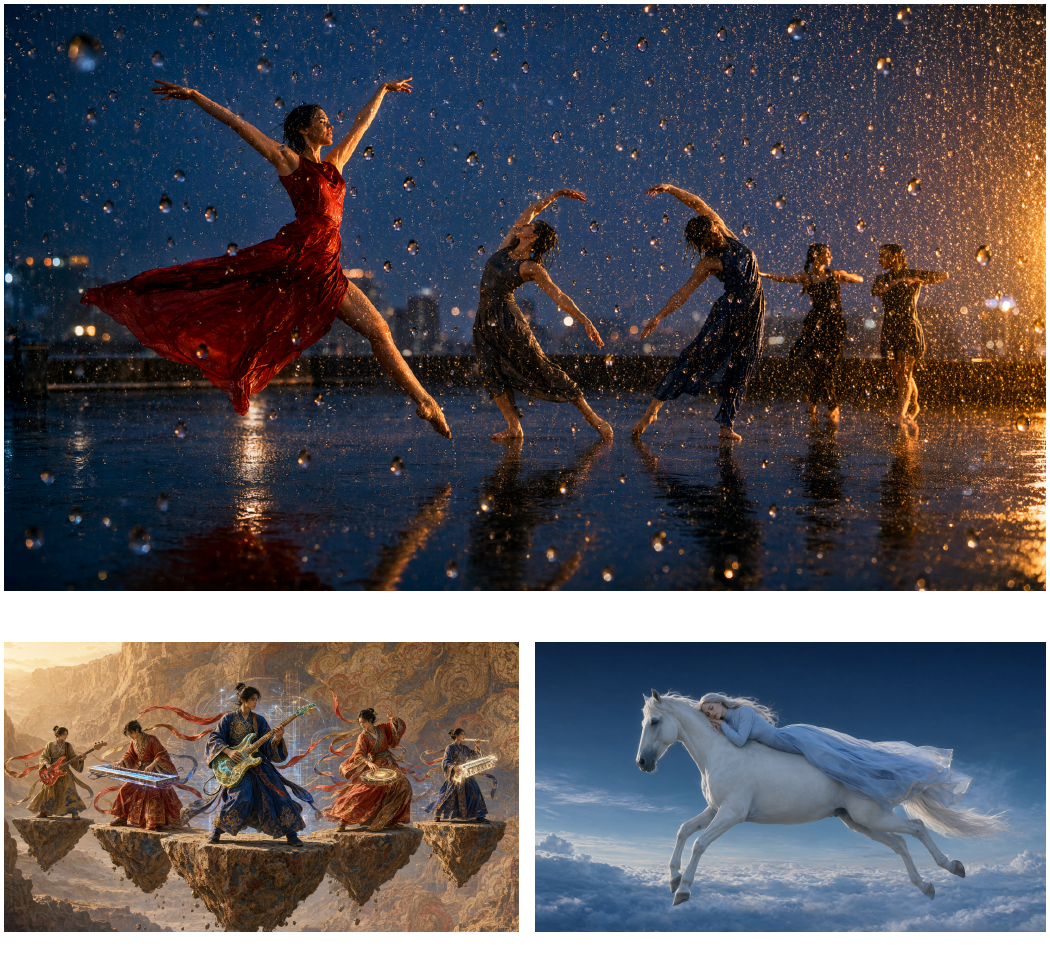}
\caption{\textbf{Qualitative gallery: motion and fantasy.} Examples of dynamic poses and imaginative visual composition.}
\label{fig:sp_g11}
\end{figure*}

\begin{figure*}[t]
\centering
\includegraphics[width=\linewidth,height=0.78\textheight,keepaspectratio]
{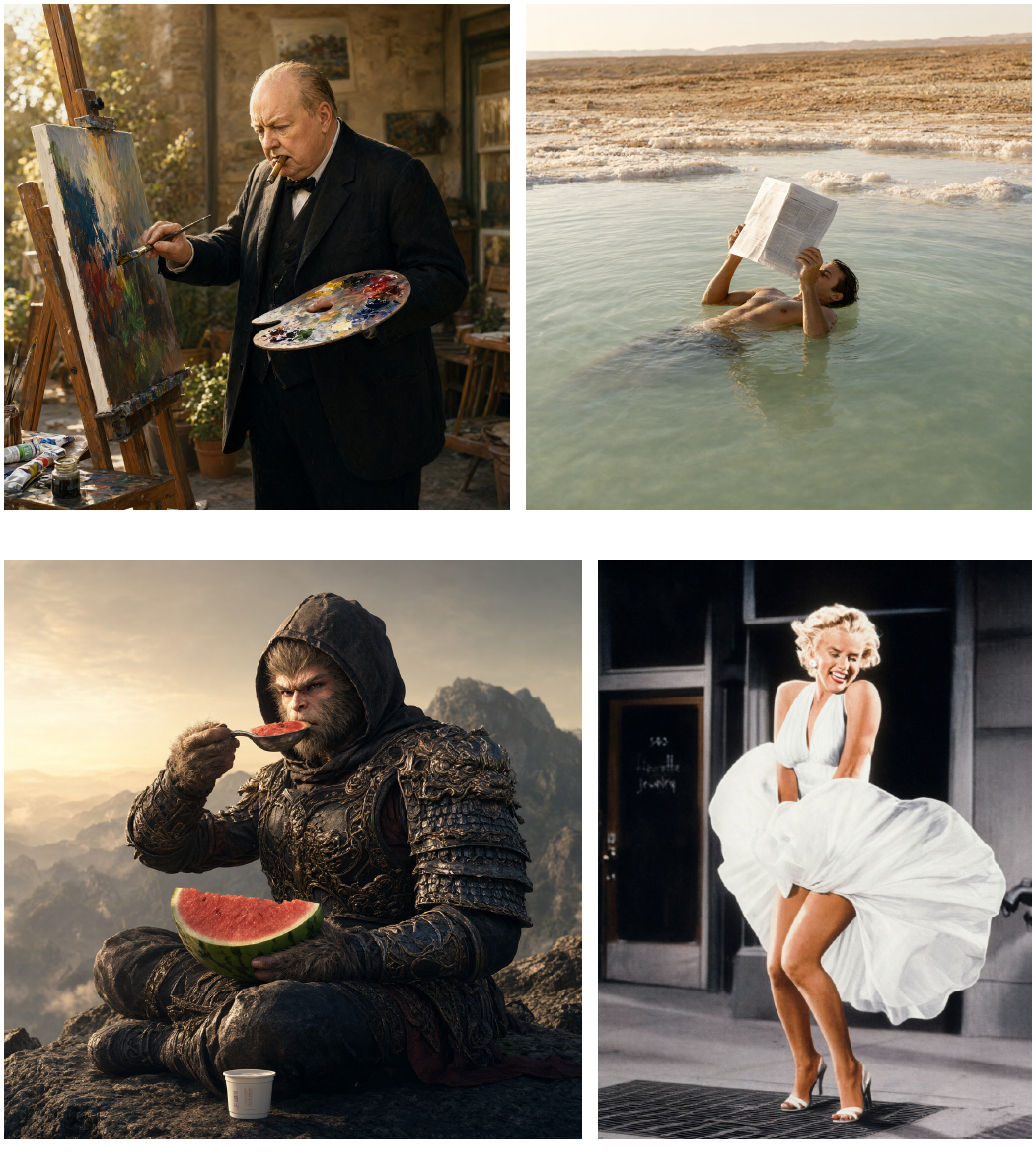}
\caption{\textbf{Qualitative gallery: people and characters.} Examples of human and character depiction across different visual settings.}
\label{fig:sp_g12}
\end{figure*}

\begin{figure*}[t]
\centering
\includegraphics[width=\linewidth,height=0.78\textheight,keepaspectratio]
{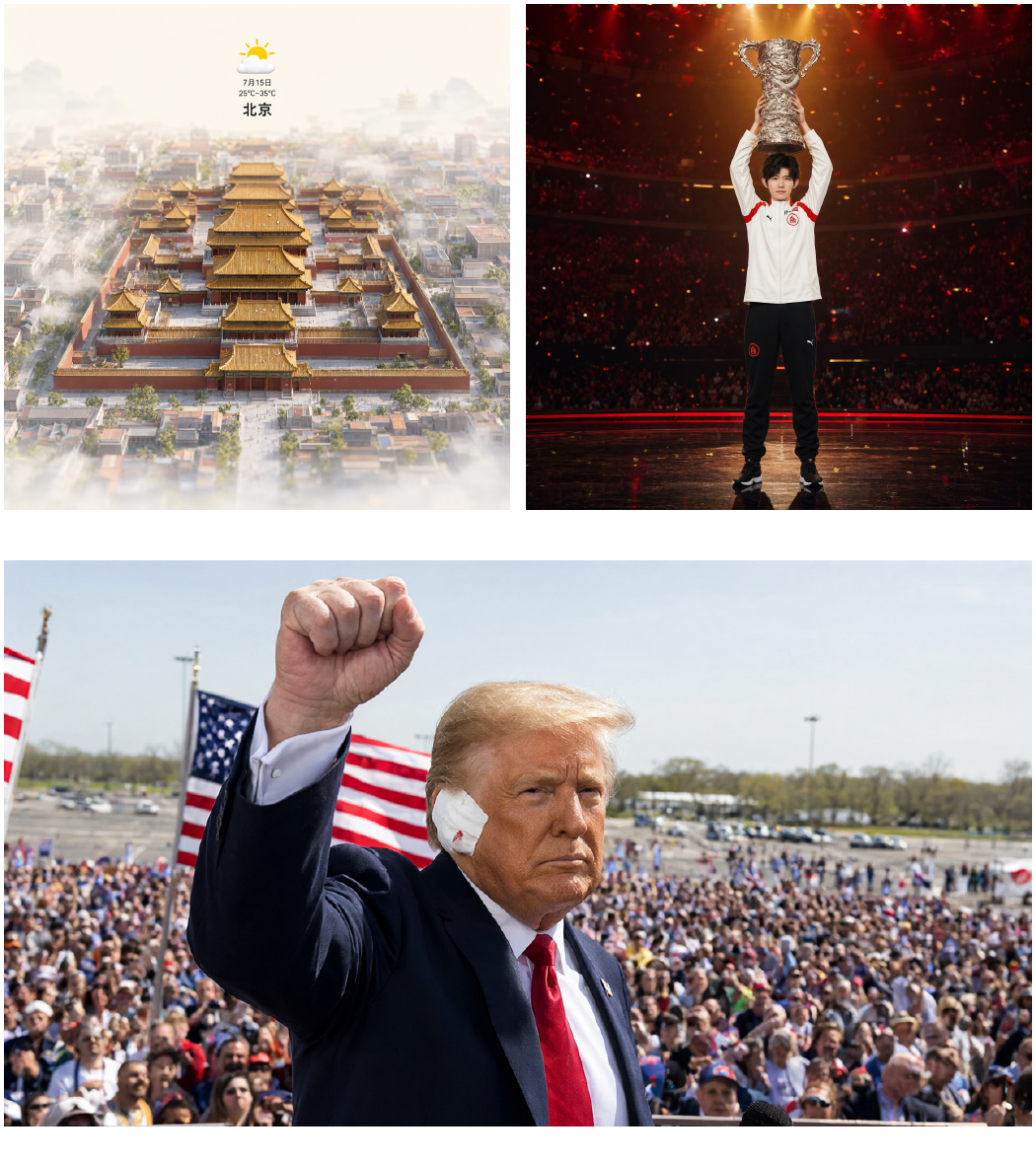}
\caption{\textbf{Qualitative gallery: culture and events.} Examples drawing on cultural themes, seasonal imagery, and public events.}
\label{fig:sp_g13}
\end{figure*}
\clearpage

\end{document}